\documentclass[a4paper,fleqn]{cas-dc}
\graphicspath{ {./figures/} }
\usepackage{amssymb}
\usepackage{multirow}
\usepackage{algorithmic}
\usepackage{array}
\usepackage{textcomp}
\usepackage{stfloats}
\usepackage{url}
\usepackage{verbatim}
\usepackage{graphicx}
\usepackage{epstopdf}
\usepackage{booktabs, siunitx, dcolumn}
\usepackage{silence}
\usepackage{bm}
\usepackage[round,authoryear]{natbib}
\usepackage{subfigure}
\usepackage{amsmath,amsfonts}
\usepackage[ruled]{algorithm2e}
\usepackage{soul}
\usepackage{bm}
\usepackage{color, xcolor}
\usepackage{times}
\usepackage{epsfig}
\usepackage{tabulary}
\usepackage{tabularx}
\usepackage{enumitem}
\usepackage{comment}
\usepackage{amssymb,amsfonts} % define this before the line numbering.
\usepackage{nicefrac}       % compact symbols for 1/2, etc.
\usepackage{microtype}      % microtypography
\usepackage{wrapfig}
\usepackage[export]{adjustbox}
\usepackage[percent]{overpic}

\soulregister \cite7
\soulregister \citep7
\soulregister\ref7

\def\tsc#1{\csdef{#1}{\textsc{\lowercase{#1}}\xspace}}
\tsc{WGM}
\tsc{QE}
\begin{document}
\let\WriteBookmarks\relax
\def\floatpagepagefraction{1}
\def\textpagefraction{.001}

%\shorttitle{C. Zhang et~al./ Expert Systems with Applications}
%\shortauthors{zcy et~al.}

\title [mode = title]{Cerberus: Attribute-based Person Re-identification Using Semantic IDs}

% \tnotemark[<tnote number>] 


\author[1]{Chanho Eom}
\ead{cheom@cau.ac.kr}

\author[2]{Geon Lee}
\ead{geon.lee@yonsei.ac.kr}

\author[3]{Kyunghwan Cho}
\ead{kyunghwan.cho@hyundai.com}

\author[3]{Hyeonseok Jung}
\ead{hyunsukdn@hyundai.com}

\author[3]{Moonsub Jin}
\ead{jinms@hyundai.com}

\author[2]{Bumsub Ham}
\cormark[1]
\ead{bumsub.ham@yonsei.ac.kr}

\affiliation[1]{organization={Graduate School of Advanced Imaging Science, Multimedia and Film, Chung-Ang University},
            city={Seoul},
            postcode={03722},
            country={South Korea}}
\affiliation[2]{organization={School of Electrical and Electronic Engineering, Yonsei University},
            city={Seoul},
            postcode={03722},
            country={South Korea}}
\affiliation[3]{organization={Robotics LAB, Hyundai Motor Company},
            city={Uiwang-si, Gyeonggi-do},
            postcode={16082},
            country={South Korea}}
\cortext[1]{Corresponding author.}

\begin{abstract}
We introduce a new framework, dubbed \emph{Cerberus}, for attribute-based person re-identification (reID). Our approach leverages person attribute labels to learn local and global person representations that encode specific traits, such as gender and clothing style. To achieve this, we define semantic IDs (SIDs) by combining attribute labels, and use a semantic guidance loss to align the person representations with the prototypical features of corresponding SIDs, encouraging the representations to encode the relevant semantics. Simultaneously, we enforce the representations of the same person to be embedded closely, enabling recognizing subtle differences in appearance to discriminate persons sharing the same attribute labels. To increase the generalization ability on unseen data, we also propose a regularization method that takes advantage of the relationships between SID prototypes. Our framework performs individual comparisons of local and global person representations between query and gallery images for attribute-based reID. By exploiting the SID prototypes aligned with the corresponding representations, it can also perform person attribute recognition (PAR) and attribute-based person search (APS) without bells and whistles. Experimental results on standard benchmarks on attribute-based person reID, Market-1501 and DukeMTMC, demonstrate the superiority of our model compared to the state of the art.
\end{abstract}


\maketitle

\section{Introduction}\label{sec:introduction}
	The goal of person re-identification (reID) is to retrieve images of the same person from a collection of gallery images across multiple cameras. Recently, it has obtained increasing attention due to its great potential in many real-world applications such as video surveillance for finding criminals or missing persons~\citep{bi2024appearance,fu2024adaptive,du2024contrastive}. Person reID is particularly challenging as 1)~the same person looks different depending on camera angles, postures, and/or lighting conditions, and 2) different persons look similar to each other, if they take similar postures or wear similar clothes. Moreover, person reID assumes a zero-shot setting, that is, person ID labels for training and test samples do not overlap. Accordingly, learning an embedding space that discriminates visually similar persons and generalizes well on unseen data is a key factor for improving performance of person reID. In the past few years, person reID methods have achieved significant advances using an attention mechanism~\citep{liu2017hydraplus,li2018harmonious,chen2019abd,zhang2020relation,chen2020salience,li2021diverse}, human pose estimators~\citep{su2017pose,suh2018part}, or generative adversarial networks~\citep{eom2019learning,zheng2019joint}. However, they still have difficulty in distinguishing persons having similar characteristics such as clothing colors. 
	
	\begin{figure*}
			\centering
			\renewcommand*{\thesubfigure}{}
			\subfigure[(a)]{
				\begin{minipage}[t]{0.35\linewidth}
				\raisebox{0.13\height}{
					\includegraphics[width=0.98\linewidth]{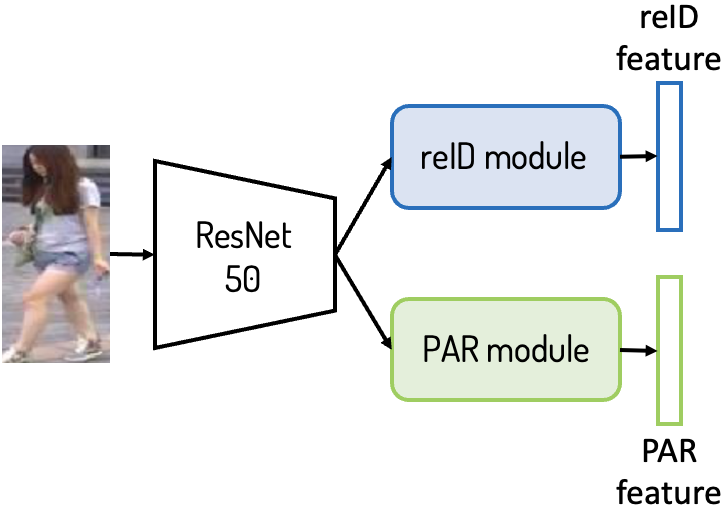}
					}
				\end{minipage}
			}
			\hspace{+0.1cm}
			\subfigure[(b)]{
				\begin{minipage}[t]{0.18\linewidth}
					\begin{overpic}[width=\linewidth]{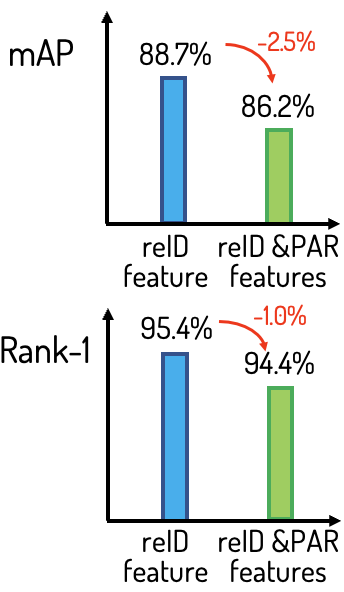}
					\end{overpic}
				\end{minipage}
			}
			\hspace{+0.3cm}
			\subfigure[(c)]{
				\begin{minipage}[t]{0.25\linewidth}
				\raisebox{0.1\height}{
					\includegraphics[width=0.98\linewidth]{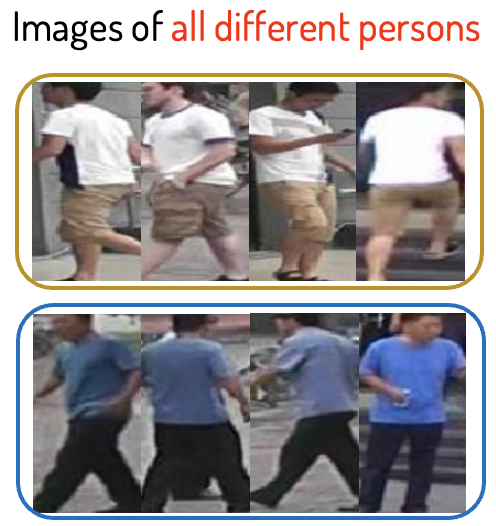}
					}
				\end{minipage}
			}
		\caption{(a) A visualization of a network architecture for existing attribute-based reID methods~\citep{liu2018ca3net,han2018attribute,tay2019aanet}. It exploits a ResNet-50~\citep{he2016deep} cropped at $\texttt{conv4-1}$ as a backbone network, and has two branches on top of that to extract features for classifying person ID and attribute labels,~\emph{i.e.},~reID and PAR features, respectively. (b) Quantitative comparisons of features for vanilla reID and attributed-based reID on Market-1501~\citep{zheng2015scalable}. Concatenating the features from both branches for a person representation rather degrades the reID performance, compared to the case that uses the reID feature alone, due to the conflicting goals between reID and PAR. (c) Examples of different persons sharing the same person attributes,~\emph{e.g.},~clothing color or gender.~(Best viewed in color.)}
	    \label{fig:teaser}
	\end{figure*}
	
	Attribute-based reID methods~\citep{lin2019improving,liu2018ca3net,han2018attribute,tay2019aanet,li2020attributes,nguyen2021graph} have been introduced that exploit person attributes as auxiliary semantic cues for reID. Complementary to ID labels, attributes provide crucial clues regarding human characteristics~(\emph{e.g.},~age, gender, hair length) that are useful for learning subtle differences between persons. In general, existing attribute-based reID methods~\citep{liu2018ca3net,han2018attribute,tay2019aanet} add an additional network for person attribute recognition (PAR) in parallel with a general reID network, and concatenate features from both networks for person representations~(Fig.~\ref{fig:teaser}(a)). However, we have found that directly using features from the PAR network as person representations rather degrades the reID performance~(Fig.~\ref{fig:teaser}(b)). We believe that this is because of the conflicting goals between PAR and reID: The crucial key for improving the reID performance is to distinguish the differences between multiple identities, even though they share the same attributes,~\emph{e.g.},~outfits or gender~(Fig.~\ref{fig:teaser}(c)). PAR, however, aims at learning visual commonness between persons sharing the same attribute labels. Consequently, features from the PAR network tend to be similar if persons share the same personal traits, and this makes person representations of similarly looking persons to be embedded closely, which degenerates the reID performance.
	
	In this paper, we present a novel framework for person reID, dubbed \emph{Cerberus}, where we use person attribute labels to guide embeddings of person representations and to help our model discriminate subtle differences between persons. To this end, we categorize person attribute labels that correlate with each other into head, upper body, lower body, identity, and carryings groups. We then define semantic identities (SIDs) as every combination of person attributes in each group. For example, the lower body group includes bottom length, color, and style attributes, and each attribute has \{short, long\}, \{red, blue, black\}, and \{pants, dress\} labels, respectively. We totally have 12 SIDs in the group,~\emph{e.g.},~`short red dress' or `long black pants'. We learn prototypical features of each SID, and use them to guide embeddings of person representations. Specifically, we extract multiple person representations, each of which describes personal traits related to head, upper body, lower body, identity, and carryings of persons. To learn person representations and SID prototypes, we introduce a semantic guidance loss that pulls representations of persons with the same SID close to the corresponding SID prototypes. For instance, we align partial representations of persons wearing,~\emph{e.g.},~`white short T-shirt' with the corresponding SID prototype for the upper body. We repeat this with the prototypes of other embedding spaces~(\emph{e.g.},~head, lower body), encouraging each representation to encode semantic information of the corresponding SID. At the same time, we enforce our model to discriminate representations of different persons but having the same SID, allowing our model to distinguish visually similar persons even with subtle appearance differences~(Fig.~\ref{fig:teaser}(c)). Note that there could be unseen SIDs at training time, since person reID is a zero-shot retrieval task. In this case, the prototypes of unseen SIDs might not be learned. To mitigate this, we also propose a regularization method that leverages relations between SID prototypes to estimate prototypes of unseen SIDs, improving the generalization performance of our model.

	During evaluation, we compute the similarity of two persons using the representations for the head, upper body, lower body, identity, and carryings individually, and average them to obtain a similarity score. We note that our framework can also perform a PAR task~(\emph{i.e.}~recognizing person attributes of a given person) and attribute-based person search~(APS) task~(\emph{i.e.}~finding pedestrians with text-based queries), without bells and whistles. This is because our framework learns a joint visual-semantic embedding space, where person representations are aligned with the corresponding SID prototypes. SID prototypes can thus be used as nearest-neighbor classifiers, and we can recognize attributes of a given person by finding the SID prototypes that give the highest matching scores with the person representations for PAR. Also, we can replace query attributes with the corresponding SID prototypes, and use them for computing similarities with person representations of gallery images, enabling retrieving persons without using any visual clue for APS. We can even search persons with partial text queries, since we align each partial person representation separately with the corresponding SID prototype in multiple visual-semantic embedding spaces. For example, our framework enables retrieving a man carrying a backpack without access to other information such as clothing color. To the best of our knowledge, this is the first model that can perform reID, PAR, and APS tasks without fine tuning for each task. We demonstrate the effectiveness of Cerberus on standard attribute-based reID benchmarks, Market-1501~\citep{zheng2015scalable} and DukeMTMC-reID~\citep{zheng2017unlabeled}, and show that it achieves competitive performances on all three tasks: reID, PAR, and APS.
	Our contributions can be summarized as follows:
		\begin{itemize}[leftmargin=*]
			\item[$\bullet$] We introduce a novel framework, dubbed \emph{Cerberus}, that exploits person attribute labels for learning multiple person representations, where each encodes particular traits of a given person to discriminate subtle differences between visually similar persons. This enables performing three different tasks, attributed-based reID, PAR, and (partial) APS, using a unified model.
			\item[$\bullet$] We propose a semantic guidance loss using attribute labels for guiding embeddings of person representations, and introduce a regularization method that enhances the generalization ability of our model on unseen data.
			\item[$\bullet$] Our model achieves the state of the art on standard attribute-based reID benchmarks, and also shows competitive performances on PAR and APS without any fine-tuning.
		\end{itemize}

\section{Related work}
In this section, we review representative works pertinent to ours, including general person reID, attribute-based person reID, APS and PAR.

\subsection{Person reID}
	Existing methods~\citep{wang2018learning,zhang2020relation,chen2020salience,li2021diverse} typically combine global and local features for robust person representations, and they can be categorized depending on how they extract local features that encode part-level person features. Attention techniques are widely adopted to extract local features focusing on salient regions,~\emph{e.g.},~body parts~\citep{liu2017hydraplus,li2018harmonious,zhang2020relation,chen2020salience,li2021diverse}. Specifically, HydraPlus-Net~\citep{liu2017hydraplus} and HA-CNN~\citep{li2018harmonious} insert attention modules into multiple levels of a backbone network, aggregating local features from low- to semantic-levels. Inspired by the work of~\cite{wang2018non}, RGA-SC~\citep{zhang2020relation} and SCSN~\citep{chen2020salience} adopt a self-attention mechanism to capture salient features from non-local regions. These methods learn attention maps in a weakly-supervised manner~(\emph{i.e.},~trained with ID labels only), and the obtained attention maps tend to focus only on the most informative region in an image, missing other diverse cues. To overcome the limitation, recent methods~\citep{zhao2017spindle,suh2018part,guo2019beyond} propose to predict body parts using,~\emph{e.g.},~body parsing models~\citep{liang2018look}. SpindleNet~\citep{zhao2017spindle} decomposes a person image into local regions,~\emph{e.g.},~head-shoulder or arm regions, and aggregates features from each local region in a coarse-to-fine manner using a tree-structured network. P$^2$-Net~\citep{guo2019beyond} extends this idea by exploiting self-attention modules to predict masks for non-human parts,~\emph{e.g.},~umbrella or bag, with an assumption that there could be useful cues to identify persons which are not related to predefined body parts. Although this approach enables providing person representations robust against deformations of body parts, it requires extra datasets with,~\emph{e.g.},~body segmentation labels (which are labor-intensive to obtain). A uniform partition strategy has recently been introduced dividing an image at equal intervals and extracting features from each partition~\citep{sun2018beyond,wang2018learning}. This approach gives large performance gains, but it is prone to spatial misalignments between body parts across images, due to the localization error caused by off-the-shelf object detectors~\citep{felzenszwalb2008discriminatively,ren2015faster}. We also extract global and local person representations. However, unlike existing methods, we explicitly guide each person representation to encode particular characteristics of a given person. This allows our model to distinguish persons sharing similar personal traits by focusing on details in such characteristics, which is essential for boosting the reID performance.

\subsection{Attribute-based person reID}
	Recent reID methods propose to exploit person attributes as auxiliary semantic cues for identifying persons. APR~\citep{lin2019improving} adopts a two-stream network, where each network is trained for reID and PAR, respectively. APR concatenates the features from each network, and uses them for person representations. Adopting APR, AANet~\citep{tay2019aanet} further leverages person attribute labels for localizing body parts, and CA$^3$Net~\citep{liu2018ca3net} propose to predict attributes sequentially using LSTM~\citep{hochreiter1997long}. AttKGCN~\citep{jiang2019attkgcn} and GPS~\citep{nguyen2021graph} have found that there exist correlations between attributes, and propose to use GCNs~\citep{kipf2017semi} to encode the correlations. Aforementioned methods, however, do not consider our observation in Fig.~\ref{fig:teaser}(b) that directly exploiting the features from attribute networks degenerates the reID performance, especially when matching persons who share the same attributes. APDR~\citep{li2020attributes} instead uses attribute labels to refine person representations. However, it requires a multi-stage training scheme, and features from the athtribute network are still integrated into person representations via fully-connected layers. Different from existing attribute-based reID methods, we do not simply use person attribute labels to train a PAR network and/or attention modules, but leverage them in order for disentangling person representations into multiple partial representations and learning minor differences between persons who share the same attributes. This does not cause the conflicting goal problem between reID and PAR shown in Fig.~\ref{fig:teaser}(b), and allows effectively improving the reID performance using person attribute labels. Moreover, our model can perform attribute-based reID, PAR, and APS using a unified model without any fine-tuning.

\subsection{Attribute-based person search}
	Person reID assumes that there is at least one query image of a person of interest, which is not always valid in real-world scenarios. To relax this assumption, lots of methods~\citep{chen2018improving,wang2020vitaa,zhao2021weakly} propose to leverage verbal descriptions of witnesses as a query for finding persons. However, they suffer from the inherent ambiguity in natural language,~\emph{i.e.},~there could be lots of possible descriptions explaining the same person. To handle this, recent methods~\citep{yin2018adversarial,cao2020symbiotic,jeong2021asmr} use a predefined set of person attributes as a query instead. These methods learn a joint visual-text embedding space, where image representations are aligned with corresponding attributes embeddings. AAIPR~\citep{yin2018adversarial} and SAL~\citep{cao2020symbiotic} adopt adversarial learning techniques to reduce the modality gap between images and attributes. ASMR~\citep{jeong2021asmr} introduces a regularization method that considers semantic distances to embed attribute representations. The limitation of these methods is that they handle global alignments between embeddings of images and attributes only. AIHM~\citep{dong2019person} proposes to align visual-text embeddings at multiple hierarchical levels, which enables local matchings between visual and text features. Similarly, we learn visual-text alignments in multiple embedding spaces. However, different from AIHM, we do not deploy extra matching networks for aligning the embeddings, which is computationally heavy in inference. Also, for the first time, our framework can search persons with partial text queries, since we learn multiple independent embedding spaces, where each encodes different semantics for corresponding grouped attribute labels,~\emph{i.e.},~SIDs.

\subsection{Person attribute recognition}
	Early methods~\citep{li2015multi,sudowe2015person} treat each person attribute independently, and train individual classifiers for each attribute using a binary cross-entropy (BCE) loss. Recently, attention mechanisms are adopted to focus on attribute-related regions in a person image. LG-Net~\citep{liu2018localization} leverages CAM~\citep{zhou2016learning} of each classifier to extract attribute-wise local features, and ALM~\citep{tang2019improving} employs a feature-pyramid attention modules~\citep{lin2017feature} to discover the most discriminative regions at multiple levels, enhancing the attribute localization accuracy. The aforementioned methods however neglect the relationships between person attributes,~\emph{e.g.},~a person wearing a pink dress with long hair is likely to be female. To address this issue, JRL~\citep{wang2017attribute} sequentially predicts attributes using LSTM~\citep{hochreiter1997long}, exploring sequential correlations between the attributes, but this requires a predefined prediction order. Recently, graph-based methods~\citep{tan2020relation,li2019pedestrian} are introduced that use GCNs~\citep{welling2017semi} to model inter-attribute correlations. JLAC~\citep{tan2020relation} employs GCNs to extract attribute-specific features and to explore the contextual relations between local regions of a given image. JVSR~\citep{li2019pedestrian} additionally leverages a human parsing network to consider spatial contexts between body parts for PAR. Our model can also perform PAR as a by-product of jointly learning person representations with SID prototypes. That is, we leverage learned SID prototypes as nearest-neighbor classifiers, and recognize the set of attributes of a given person. However, since the proposed framework is not designed for PAR but for reID, it does not exploit specialized components for PAR such as a BCE loss or GCNs.

\section{Approach}\label{sec:proposed}
	\vspace{-0.1cm}
In this section, we first describe our framework for attribute-based person reID~(Section~\ref{subsec:overview}), and then provide detailed explanations for training losses~(Section~\ref{subsec:training}).
	
	\begin{figure*}[t]
	    \centering
	    \includegraphics[width=\linewidth]{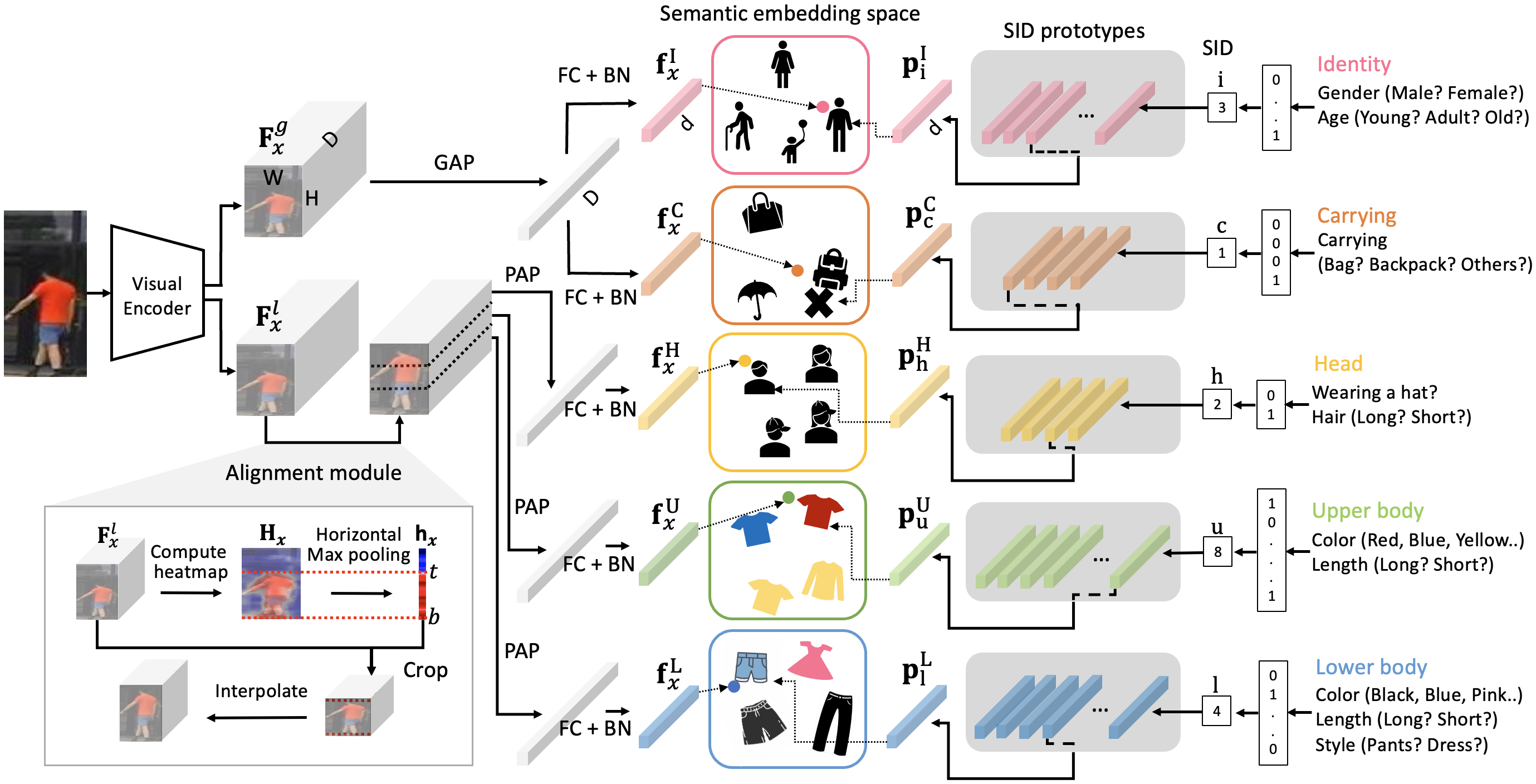}
	    \caption{An overview of \emph{Cerberus}. We extract global and local feature maps, denoted by $\mathbf{F}_x^{g}$ and $\mathbf{F}_x^{l}$, respectively, from a given image. We then apply global average pooling (GAP) to the global feature map~$\mathbf{F}_x^{g}$, and use fully connected (FC) and batch-norm (BN) layers to obtain representations for identity~($\mathbf{f}_x^\mathrm{I}$) and carrying~($\mathbf{f}_x^\mathrm{C}$), where the size of each representation is $d$. Similarly, we incorporate a part average pooling (PAP) layer, followed by a series of fully connected (FC) and batch normalization (BN) layers, on the local feature map~$\mathbf{F}_x^{l}$ to extract representations for the head, upper body, and lower body, denoted by $\mathbf{f}_x^\mathrm{H}$, $\mathbf{f}_x^\mathrm{U}$, and $\mathbf{f}_x^\mathrm{L}$, respectively, from the top, middle, and bottom parts of the image. Note that, for the local feature map $\mathbf{F}_x^{l}$, we insert an alignment module that estimates the region, where a person is likely to exist. We define SIDs, and learn corresponding prototypical features~($\mathbf{p}^\mathrm{I}_\mathrm{i}$, $\mathbf{p}^\mathrm{C}_\mathrm{c}$, $\mathbf{p}^\mathrm{H}_\mathrm{h}$, $\mathbf{p}^\mathrm{U}_\mathrm{u}$, and $\mathbf{p}^\mathrm{L}_\mathrm{l}$), which are used to guide embeddings of person representations. See the text for more details.~(Best viewed in color.)}
	    \label{fig:overview}
	\end{figure*}
	
	\subsection{Architecture}\label{subsec:overview}
	 We represent person images using multiple partial representations that encode features related to head, upper body, lower body, identity, and carryings, to discriminate the query person from others. To this end, we leverage person attribute labels to disentangle person representations into multiple partial representations and to guide the embeddings of each representation. Specifically, we categorize attribute labels into head, upper body, lower body, identity, and carryings groups, and define SIDs by combining the attributes in the particular group. We then learn the prototypical features of each SID, and use them to embed partial representations of the persons having the same SID,~\emph{i.e.},~visually similar persons, nearly in the embedding space. This enables the representations to encode corresponding semantics of SIDs. Simultaneously, we encourage the representations of the same person to form a compact cluster so that they can be distinguished from the representations of others, learning subtle appearance differences between persons sharing similar attributes. We regularize SID prototypes using semantic relations to improve the generalization ability of our method. Our model is trained end-to-end for person reID. After training, it can also be used for APS and PAR without additional fine-tuning for each task. We show an overview of \emph{Cerberus} in Fig.~\ref{fig:overview}.

	\begin{table*}[t]
		\caption{Examples of grouped attribute labels for Market-1501~\citep{zheng2015scalable} and DukeMTMC-reID~\citep{zheng2017unlabeled}.}
		\label{tab:attribute_group}
		\centering
		\resizebox{0.8\linewidth}{!}{
			\begin{tabular}{p{2cm} p{6.2cm} p{4.5cm}}
			\toprule
			Group & Market-1501 & DukeMTMC-reID \\
			\midrule
			Head 		& hat, hair length							& hat \\
			Upper body 	& top color, sleeve length 					& top color, sleeve length \\
			Lower body 	& bottom color, bottom length, bottom style 	& bottom color, shoe color, boots \\
			Identity 	& gender, age 								& gender \\
			Carrying 	& backpack, bag, handbag 					& backpack, bag, handbag \\
			\bottomrule
			\end{tabular}
		}
	\end{table*}

		\subsubsection{Person representations}
		We describe a person image using multiple partial representations that encode personal traits in head, upper body, lower body, identity, and carryings of a given person. To this end, we extract two feature maps, $\mathbf{F}_x^{g}, \mathbf{F}_x^{l} \in \mathbb{R}^{H \times W \times D }$, from the person image to extract global and local person features, respectively, where $H$, $W$, and $D$ are height, width and channel depth of the feature maps, respectively. We then obtain the partial representations by applying pooling, fully-connected (FC), and batch norm (BN)~\citep{ioffe2015batch} layers. To be specific, from the local feature map~$\mathbf{F}_x^{l}$, we extract representations for head, upper body, and lower body, denoted by $\mathbf{f}_x^\mathrm{H}$,~$\mathbf{f}_x^\mathrm{U}$, and~$\mathbf{f}_x^\mathrm{L}$, respectively, which are associated with particular local regions in the person image. On the other hand, representations for identity and carrying, $\mathbf{f}_x^\mathrm{I}$ and $\mathbf{f}_x^\mathrm{C}$, are extracted from the global feature map~$\mathbf{F}_x^{g}$, since relevant regions for specifying identity and carrying may not be fixed within the image. To exploit the prior knowledge that head, upper body, and lower body are probably located in the top, middle, and bottom of an image, respectively, we apply a part average pooling (PAP) method for the local features, while a global average pooling (GAP) method is used for the global ones. Note that, considering a person is often located only at a certain part of the image due to the localization error of off-the-shelf person detectors, we use a simple alignment module that estimates the region, where a person is likely to exist from an image. Specifically, we obtain a heat map $\mathbf{H}_x \in \mathbb{R}^{H \times W}$ by computing the magnitude of the local feature map~$\mathbf{F}_x^{l}$ in each spatial position $p$,~\emph{i.e.},~$\mathbf{H}_x(p)= \big\lvert \mathbf{F}_x^{l}(p) \big\rvert_2$. We then apply a max-pooling operator on the heat map~$\mathbf{H}_x$ along the horizontal direction, and obtain $\mathbf{h}_x \in \mathbb{R}^{H}$. We find the smallest and largest indexes, $t$ and $b$, where $\mathbf{h}_x$ is larger than a pre-defined threshold $\sigma$. We then discard features from the regions outside of the range t and b, assuming that the magnitude of a feature extracted from a human body part is much larger than others, which is reasonable because the model tends to focus more on the body part as training progresses~\citep{wang2018learning,zheng2019re}. We resize the cropped feature map into the original size via bilinear interpolation. Note that, compared to previous methods~\citep{su2017pose,li2017learning,li2018harmonious,li2020attributes} that use STN~\citep{jaderberg2015spatial} or attention modules for localizing human body parts, this alignment module does not require any learnable parameters.
				
	\begin{figure}[t]
	    \centering
	    \includegraphics[width=0.9\linewidth]{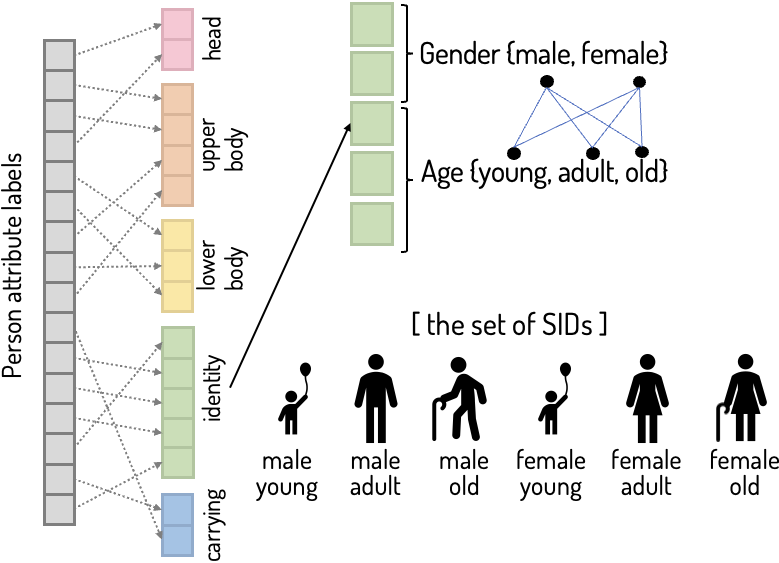}
	    \caption{Illustrations of constructing the set of semantic IDs. See the text for more details.~(Best viewed in color.)}
	    \label{fig:sid_illust}
	\end{figure}

		\subsubsection{SID prototypes} 
		We show in Fig.~\ref{fig:sid_illust} a process of constructing SIDs. We take attribute labels of a person image, which are represented as a binary vector, where each dimension indicates the presence or absence of a certain attribute with 1 or 0, respectively. Motivated by \cite{zhao2018grouping,li2020attributes,nguyen2021graph}, we divide the labels into disjoint groups that are necessary for describing person. Each group contains labels related to the head, upper body, lower body, identity, and carrying of persons, respectively. Note that, as shown in Table~\ref{tab:attribute_group}, regardless of the dataset having different attribute labels, mapping specific attribute labels to their corresponding groups enables easy extension to each dataset. We combine attributes in each group, and define the sets of SIDs for head, upper body, lower body, identity, and carrying, denoted by~$\mathcal{S}^\mathrm{H}$, $\mathcal{S}^\mathrm{U}$, $\mathcal{S}^\mathrm{L}$, $\mathcal{S}^\mathrm{I}$, and $\mathcal{S}^\mathrm{C}$, respectively. For instance, the identity group for Market-1501~\citep{zheng2015scalable} contains age and gender attributes, where each attribute has \{young, adult, old\} and \{male, female\} labels, respectively. Consequentially, there are 6 SIDs,~\emph{e.g.},~`young male' or `adult female' in the identity group. We denote by $\mathrm{h}$, $\mathrm{u}$, $\mathrm{l}$, $\mathrm{i}$, and $\mathrm{c}$ SIDs of a given image for head, upper body, lower body, identity, and carrying groups, respectively,~\emph{e.g.},~2, 8, 4, 3 and 1 in Fig.~\ref{fig:overview}. Similarly, corresponding prototypes are denoted by $\mathbf{p}^\mathrm{H}_\mathrm{h}$, $\mathbf{p}^\mathrm{U}_\mathrm{u}$, $\mathbf{p}^\mathrm{L}_\mathrm{l}$, $\mathbf{p}^\mathrm{I}_\mathrm{i}$, and $\mathbf{p}^\mathrm{C}_\mathrm{c} \in \mathbb{R}^d$.
%		Motivated by \cite{zhao2018grouping,li2020attributes,nguyen2021graph}, we divide these labels into disjoint subsets, where each subset contains labels related to head, upper body, lower body, identity, and carrying of persons, respectively (See Table~\ref{tab:attribute_group} for example).

	\begin{figure*}
		\centering
		\renewcommand*{\thesubfigure}{}
		\subfigure[(a)]{
			\begin{minipage}[t]{0.32\linewidth}
				\includegraphics[width=0.98\linewidth]{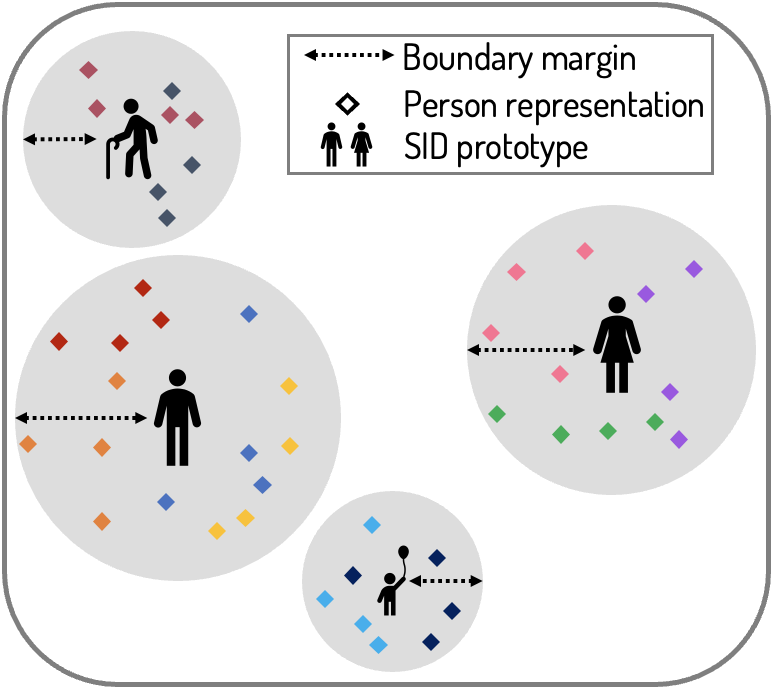}
			\end{minipage}
		}
		\subfigure[(b)]{
			\begin{minipage}[t]{0.32\linewidth}
				\includegraphics[width=0.98\linewidth]{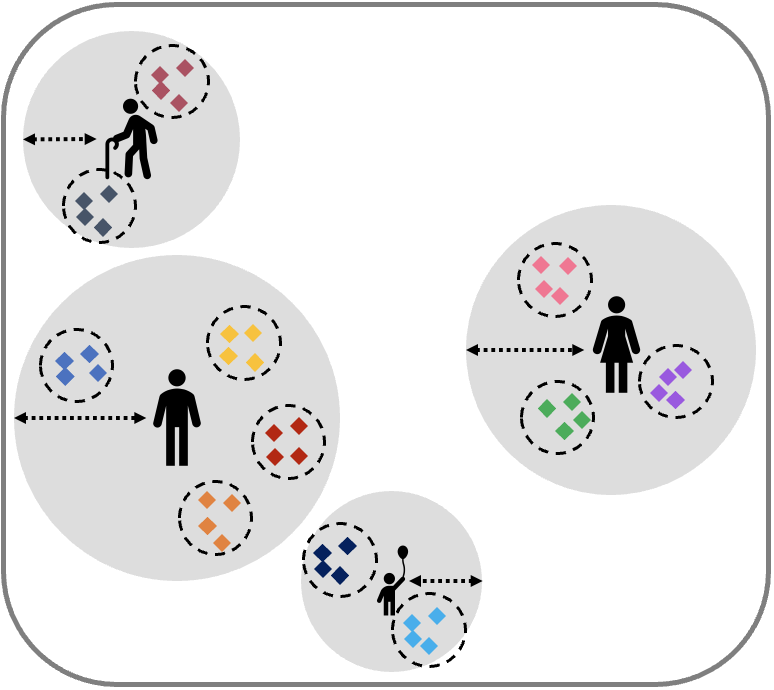}
			\end{minipage}
		}
		\subfigure[(c)]{
			\begin{minipage}[t]{0.32\linewidth}
				\includegraphics[width=0.98\linewidth]{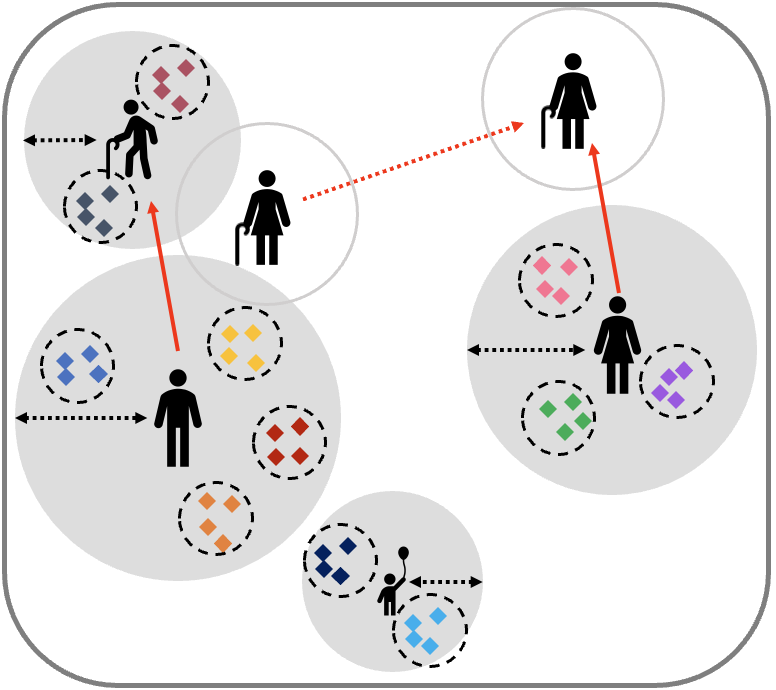}
			\end{minipage}
		}
		\caption{Illustrations of the embedding spaces in our model. (a) The semantic guidance term encourages the representations of persons belonging to the same SID to be grouped close to the corresponding SID prototype. (b) The identification term enables the representations of the same person to form clusters. Accordingly, the two terms allow us to differentiate subtle differences between SIDs and ID labels. (c) We constraint the SID prototypes by their semantic relations, enabling estimating prototypes of unseen SIDs. For example, if there is no person belonging to `old female' in the training data, its SID prototype may be positioned incorrectly in the embedding space. Using the regularization loss, we encourage the SID prototype for `old female' to be placed near `adult female', reflecting the relationship between the prototypes for `adult male' and `old male' (represented by the red dotted line). The red solid lines indicate the residual vectors as defined in Eq.~\eqref{eq:residual_vector}. The points with the same color indicate that they correspond to the same identity. See the text for more details.~(Best viewed in color.)}
	    \label{fig:sid_concept}
	\end{figure*}

	\subsection{Training loss} \label{subsec:training}
		The learning objective of our model is defined as follows:
			\begin{equation}
				\mathcal{L} = \mathcal{L}_{embed} + \lambda_{reg}\mathcal{L}_{reg},
			\end{equation}
			where $\mathcal{L}_{embed}$ and $\mathcal{L}_{reg}$ are embedding and regularization terms, respectively, and $\lambda_{reg}$ is a balance parameter. We provide details of each loss in the following.
				
		\subsubsection{Embedding loss} The embedding loss consists of two components:
			\begin{equation}
				\mathcal{L}_{embed} = \lambda_{sem}\mathcal{L}_{sem} + \lambda_{id}\mathcal{L}_{id},
			\end{equation}
			where $\mathcal{L}_{sem}$ and $\mathcal{L}_{id}$ denote semantic guidance and identification terms, respectively, and $\lambda_{sem}$ and $\lambda_{id}$ are weighting factors for each loss.
		
		\paragraph{\textbf{Semantic guidance term.}}
			
			We extract multiple person representations that describe personal traits such as head, upper body, lower body, identity, and carrying for each individual. These person representations are then embedded in separate embedding spaces along with the corresponding SID prototypes. Namely, we align the person representations with SID prototypes in multiple embedding spaces.  For example, upper body representations of persons wearing a `short red top' are encouraged to be placed close to one another in the corresponding embedding space. This encourages the representations to encode the semantics of the corresponding SID, and allows persons who share the same semantic concept to be embedded closely. To achieve this, we define a semantic guidance loss as follows:
				\begin{equation}
					\mathcal{L}_{sem} = \frac{1}{\lvert \mathcal{P} \rvert} \sum_{(\mathrm{G},\mathrm{g}) \in \mathcal{P}} max\big( 1 - m^\mathrm{G}_\mathrm{g} - s(\mathbf{f}_x^\mathrm{G}, \mathbf{p}^\mathrm{G}_\mathrm{g}), \text{~}0 \big).
				\label{eq:l_sem}
				\end{equation}
				We denote by $\mathcal{P} = \{ (\mathrm{H},\mathrm{h}), (\mathrm{U},\mathrm{u}), (\mathrm{L},\mathrm{l}), (\mathrm{I},\mathrm{i}), (\mathrm{C},\mathrm{c}) \}$, the set of pairs, where each pair consists of an attribute group and the corresponding SID label of a given image. $s(\cdot,\cdot)$ computes cosine similarity between inputs, and $m^\mathrm{G}_\mathrm{g}$ is a boundary margin, defined as follows:
				\begin{equation}
					m^\mathrm{G}_\mathrm{g} = \log \bigg(\alpha \cdot \frac{N^\mathrm{G}_\mathrm{g}}{N} + \beta \bigg),
				\label{eq:boundary_margin}
				\end{equation}
				where $N^\mathrm{G}_\mathrm{g}$ is the number of persons belonging to the $\mathrm{g}$-th SID of the group $G$, and $N$ is the number of total persons in training data. $\alpha$ and $\beta$ are hyperparameters that control the slope and bias of the $\log$ function. The semantic guidance loss aligns the representations of head, upper body, lower body, identity, and carrying with the corresponding SID prototypes until the similarity between them exceeds $1 - m^\mathrm{G}_\mathrm{g}$. This results in person representations belonging to the same SID are closely placed in the embedding space within a certain boundary $1 - m^\mathrm{G}_\mathrm{g}$~(Fig.~\ref{fig:sid_concept}(a)).		
	
			\paragraph{\textbf{Identification term.}}
				The semantic guidance loss allows our representations to reflect the attributes of a given person. However, the subtle visual differences between persons with the same attributes remain undetected. To further discriminate between persons with the same SID, we encourage person representations to be clustered according to ID labels using the identification loss defined as follows:
				\begin{flalign}
				\label{eq:l_id}
					\mathcal{L}_{id} = \frac{1}{\lvert \mathcal{G} \rvert}
						\sum_{\mathrm{G} \in \mathcal{G}} &\bigg(-log p\big(y_x|\mathbf{f}_x^\mathrm{G} \big) \\ \nonumber
						+ &\log \Big(1+exp\big(d(\mathbf{f}_x^\mathrm{G}, \mathbf{f}_p^\mathrm{G})-d(\mathbf{f}_x^\mathrm{G}, \mathbf{f}_n^\mathrm{G})\big)\Big) \bigg),
				\end{flalign}
			 where $\mathcal{G} = \{ \mathrm{H}, \mathrm{U}, \mathrm{L}, \mathrm{I}, \mathrm{C} \}$ and $p\big(y_x|\mathbf{f}_x^\mathrm{G} \big)$ is the probability that the representation $\mathbf{f}_x^\mathrm{G}$ belongs to $y_x$, where $y_x$ is the ID label of the $x$-th image. $d(\cdot,\cdot)$ computes the Euclidean distance between inputs. $\mathbf{f}_p^\mathrm{G}$ is the person representation which has the same ID label as an anchor $\mathbf{f}_x^\mathrm{G}$, while $\mathbf{f}_n^\mathrm{G}$ is the negative one having a different ID label. The former encourages our representations to be discriminative enough for identifying person IDs, while the latter enforces intra-person distances to be smaller than inter-person distances, allowing person representations to form compact clusters based on their ID labels in the embedding space~(Fig.~\ref{fig:sid_concept}(b)). The identification term promotes our model to focus on subtle appearance differences, such as printing on T-shirts, to distinguish persons wearing similar clothes. This leads to person representations containing information about unique characteristics of a person, including head, upper body, lower body, identity, and carryings.

				To summarize, our embedding loss balances the trade-off between the semantic guidance term (Eq.~\eqref{eq:l_sem}) and the identification term (Eq.~\eqref{eq:l_id}). The semantic guidance term encourages close embedding of person representations that belong to the same SID in the semantic embedding space. The identification term, on the other hand, enforces clear separation between the representations of different persons. When the distance between a person representation and its corresponding SID prototype is smaller than the boundary margin~$m^\mathrm{G}_\mathrm{g}$, the semantic guidance term becomes zero and the identification term dominates the embedding loss, guiding our model to focus on learning unique characteristics of the person to distinguish it from others with the same attributes. For instance, persons who possess a backpack are categorized under the same SID, yet backpacks may exhibit variations in size, shape, or number of pockets, and we expect that the learned semantic embedding space will effectively differentiate such subtle differences.

				Note that the boundary margin~$m^\mathrm{G}$ in the semantic guidance loss is proportional to the number of persons belonging to the SID. The more persons belong to the same semantic concept, the greater the focus on the identification term to discover their differences.
				
	\begin{figure*}
		\centering
		\renewcommand*{\thesubfigure}{}
		\subfigure[(a) reID]{
			\begin{minipage}[t]{0.32\linewidth}
				\includegraphics[width=0.98\linewidth]{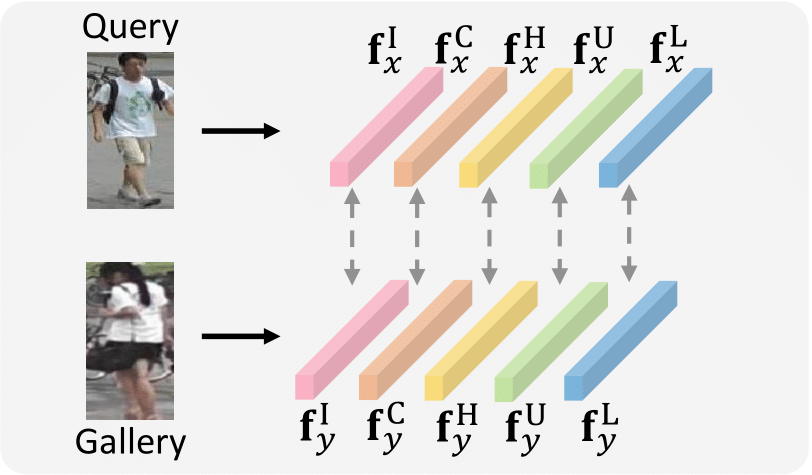}
			\end{minipage}
		}
		\subfigure[(b) APS]{
			\begin{minipage}[t]{0.32\linewidth}
				\includegraphics[width=0.98\linewidth]{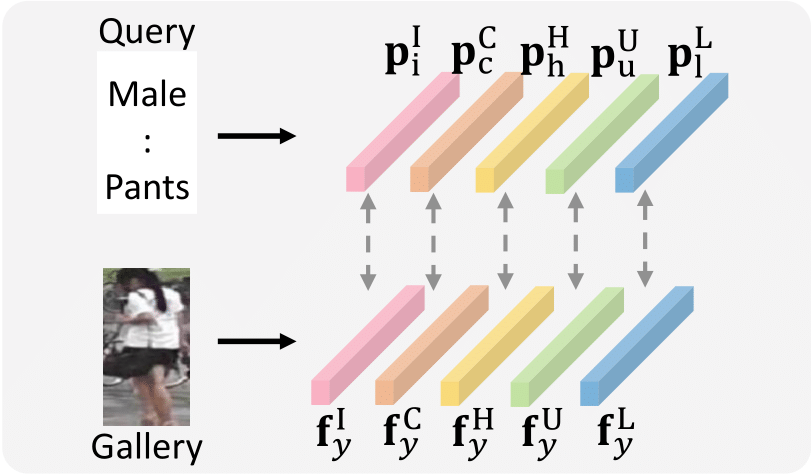}
			\end{minipage}
		}
		\subfigure[(c) PAR]{
			\begin{minipage}[t]{0.32\linewidth}
				\includegraphics[width=0.98\linewidth]{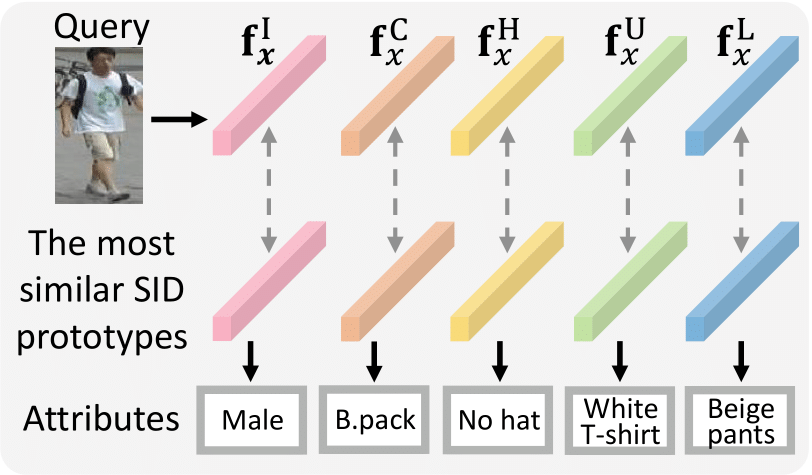}
			\end{minipage}
		}
		\caption{Illustrations of inference processes for reID, APS, and PAR. (a)~\textbf{reID}: We compare person representations of query and gallery images by computing cosine similarity between individual partial representations. (b)~\textbf{APS}: We replace query representations with SID prototypes that the query belongs to, and calculate cosine similarity with person representations of the query. (c)~\textbf{PAR}: We find SID prototypes that show the highest matching score with each partial representation of the query, and convert their SIDs into attributes.~(Best viewed in color.)}
        \label{fig:inference}
	\end{figure*}

		\subsubsection{Regularization loss} 
		
		We learn multiple visual-semantic embedding spaces, where partial representations for head, upper body, lower body, identity, and carrying are aligned with the corresponding SID prototypes in each embedding space. However, certain SID prototypes may not be trained if there are no persons with those SIDs in the training set~(\emph{e.g.}, the prototype of `old female' in the identity group as depicted in Fig.~\ref{fig:sid_concept}(c)). It is thus highly likely that corresponding person representations are placed incorrectly in the embedding space, which leads to difficulty in recognizing such persons. To mitigate this problem, we propose a regularization loss to constrain the embeddings of SID prototypes based on their relationships with one another, improving the ability of our model to infer prototypes of unseen SIDs and enhancing the generalization ability. We define the regularization term as follows:
		
			\begin{equation}
				\mathcal{L}_{reg} = \frac{1}{\lvert \mathcal{G} \rvert} \sum_{\mathrm{G} \in \mathcal{G}} \mathcal{L}_{reg}^\mathrm{G},
			\end{equation}
			where
			\begin{equation}
				\mathcal{L}_{reg}^\mathrm{G} = 
					\sum_{m=1}^{\big\lvert \mathcal{S}^\mathrm{G} \big\rvert}
					\sum_{n=1}^{\big\lvert \mathcal{S}^\mathrm{G} \big\rvert}
					\Big\lVert \mathbf{p}^\mathrm{G}_m - \mathbf{p}^\mathrm{G}_n - \mathbf{r}_{m,n} \Big\rVert^2.
			\end{equation}
			The regularization term constrains the relationship between all pairs of prototypes in a given group using a residual vector~$\mathbf{r}_{m,n}$. The residual vector $\mathbf{r}_{m,n}$ is defined as:
			\begin{equation}
				\mathbf{r}_{m,n} = \sum_{l=1}^{L^\mathrm{G}} \Big( \mathbf{v}_l \cdot \big( \mathbf{A}^\mathrm{G}_{m}(l) - \mathbf{A}^\mathrm{G}_{n}(l) \big) \Big),
				\label{eq:residual_vector}
			\end{equation}
			where $L^\mathrm{G}$ is the number of attributes that belong to the group $G$, and $\mathbf{v}_l$ is a learnable parameter of size $d$. We denote by $\mathbf{A}^\mathrm{G}_{m}$ corresponding attribute labels to $\mathbf{p}^\mathrm{G}_m$, which is a binary vector, where each dimension represents the presence or absence of a specific attribute. $\mathbf{A}^\mathrm{G}_{m}(l)$ represents the $l$-th value of $\mathbf{A}^\mathrm{G}_{m}$, suggesting that $\mathbf{A}^\mathrm{G}_{m} = 1$ if $\mathbf{p}^\mathrm{G}_m$ has the $l$-th attribute, and $\mathbf{A}^\mathrm{G}_{m} = 0$ otherwise. If a prototype pair, $\mathbf{p}^\mathrm{G}_m$ and $\mathbf{p}^\mathrm{G}_n$, share the same $l$-th attribute label, $\mathbf{A}^\mathrm{G}_{m}(l) - \mathbf{A}^\mathrm{G}_{n}(l)$ is equal to 0, otherwise, it takes a value of 1 or -1, determining the direction of $\mathbf{v}_l$. As a result, when the differences in attribute labels between the prototype pairs are the same, these pairs share the same residual vector. For instance, Fig.~\ref{fig:sid_concept}(c) shows two prototype pairs, (`adult male', `old male'), and (`adult female', `old female'). The residual vectors between the prototypes in each pair are then regularized to be the same, since the prototypes of both pairs share the same attribute labels except for the `adult/old' attribute. This enables our model to embed SID prototypes reflecting their semantic relations and to estimate the prototypes of unseen SIDs, such as `old female', thus improving the generalizability of our model.

		\subsection{Inference} 
		Our model learns multiple joint embedding spaces for attribute-based person reID, where individual partial representations are semantically aligned with corresponding SID prototypes. Using the joint embedding space and SID prototypes with a negligible memory overhead (See Section~\ref{subsec:discussion}), our model can also be used to retrieve person attribute descriptions~(\emph{i.e.},~APS) or recognize personal attributes from a given image~(\emph{i.e.},~PAR) without additional fine-tuning. We present detailed descriptions on applying our model to attribute-based person reID, APS, and PAR in the following.

		\paragraph{\textbf{Attribute-based person reID}~(Fig.~\ref{fig:inference}(a)).} Given a query image, we extract person representations, $\mathbf{f}_x^\mathrm{H}, \mathbf{f}_x^\mathrm{U}, \mathbf{f}_x^\mathrm{L}, \mathbf{f}_x^\mathrm{I}$, and $\mathbf{f}_x^\mathrm{C}$, and compare them with those of a gallery image,~\emph{i.e.},~$\mathbf{f}_y^\mathrm{H}, \mathbf{f}_y^\mathrm{U}, \mathbf{f}_y^\mathrm{L}, \mathbf{f}_y^\mathrm{I}$, and $\mathbf{f}_y^\mathrm{C}$. Specifically, we compute cosine similarity between corresponding representations, and average the similarity scores for matching. Note that, although we leverage SID prototypes to guide embeddings of person representations at training time, we do not use them at test time. 
		\paragraph{\textbf{APS}~(Fig.~\ref{fig:inference}(b)).} We represent input attribute labels with SID prototypes, $\mathbf{p}^\mathrm{H}_\mathrm{h}$, $\mathbf{p}^\mathrm{U}_\mathrm{u}$, $\mathbf{p}^\mathrm{L}_\mathrm{l}$, $\mathbf{p}^\mathrm{I}_\mathrm{i}$, and $\mathbf{p}^\mathrm{C}_\mathrm{c}$, by retrieving the prototypes that the labels belong to. We then compute cosine similarity between the SID prototypes and person representations from gallery images. Note that, since we learn partial person representations, each of which is aligned with SID prototypes in a disjoint embedding space, we can perform the APS task with a query having partial attribute labels. For instance, let us suppose that attributes related to the head and upper body of the query are missing. Then, we compute the similarity scores between SID prototypes, $\mathbf{p}^\mathrm{L}_\mathrm{l}$, $\mathbf{p}^\mathrm{I}_\mathrm{i}$, and $\mathbf{p}^\mathrm{C}_\mathrm{c}$, and gallery representations, $\mathbf{f}_y^\mathrm{L}, \mathbf{f}_y^\mathrm{I}$, and $\mathbf{f}_y^\mathrm{C}$. 
		\paragraph{\textbf{PAR}~(Fig.~\ref{fig:inference}(c)).} We use SID prototypes as nearest neighbor (NN) classifiers. To be specific, given person representations of the query image, we find the most similar SID prototype for each representation, and retrieve a corresponding SID as follows:
			\begin{equation}
				{s}_x^\mathrm{G} = \underset{k}{argmax}\text{~}s(\mathbf{f}_x^\mathrm{G}, \mathbf{p}_k^\mathrm{G}),\text{~~where~~} k \in \big\{1,...,\big\lvert \mathcal{S}^\mathrm{G} \big\rvert \big\}.
			\end{equation}
			 $\mathrm{G}$ indicates the attribute group,~\emph{i.e.},~$\mathrm{G} = \{ \mathrm{H}, \mathrm{U}, \mathrm{L}, \mathrm{I}, \mathrm{C} \}$. We then convert the retrieved SID, ${s}_x^\mathrm{G}$, into attribute labels.

\section{Experiments}

	\subsection{Experimental details} \label{subsec:experimental_details}
	
		\subsubsection{Datasets and evaluation metric} Following other attribute-based reID methods~\citep{lin2019improving,liu2018ca3net,tay2019aanet,li2020attributes,nguyen2021graph}, we evaluate our model on Market-1501~\citep{zheng2015scalable} and DukeMTMC-reID~\citep{zheng2017unlabeled}. We use person attribute labels provided by Lin~\emph{et. al.}~\citep{lin2019improving}, where 27 and 23 person attributes are annotated for Market-1501 and DukeMTMC-reID, respectively. We group the attribute labels correlated with each other as in Table~\ref{tab:attribute_group}, and define SIDs based on the combination of attributes in each group. As a result, there are 66 and 61 SIDs in train/test sets of Market-1501, while DukeMTMC-reID has 46 and 33 SIDs, respectively. Although attribute-based reID methods (including ours) additionally leverage person attribute labels together with ID labels during training, the attribute labels are very cheap and easy to collect, compared to,~\emph{e.g.},~body parts or human parsing masks that are widely adopted by reID approaches~\citep{suh2018part,guo2019beyond,liang2018look}. To be specific, Lin~\emph{et.al.}~\citep{lin2019improving} assume that personal traits would not significantly vary across cameras, and they annotate attribute labels of a single image alone for each person. As a result, 751 and 702 images are annotated for training on Market-1501 and DukeMTMC-reID, respectively, which are 5.81\% and 4.25\%, compared to the total number of training samples.
		
			Although our goal is to design an attribute-based person reID method addressing the conflicts between identifying persons and recognizing attributes, the proposed model has also an ability to handle PAR and APS. To show the effectiveness of our model on PAR and APS, we also exploit Market-1501~\citep{zheng2015scalable} and DukeMTMC-reID~\citep{zheng2017unlabeled}. Note that we would not use datasets specially designed for PAR and APS tasks,~\emph{e.g.},~PETA~\citep{deng2014pedestrian} or RAP~\citep{li2016richly}. Since they do not provide person ID labels and/or the number of person images of the same ID across cameras is not sufficient, we could not train our model designed for attribute-based person reID. 
		
			We measure the performance of reID and APS by computing mean average precision~(mAP) and rank-1 accuracy. For PAR, we compute the classification accuracy for each attribute and report the mean accuracy~(mA).
	
	\begin{table*}[!t]
		\caption{Quantitative comparisons with state-of-the-art methods for (attribute-based)~reID on Market-1501~\citep{zheng2015scalable} and DukeMTMC-reID~\citep{zheng2017unlabeled} in terms of rank-1 accuracy(\%) and mAP(\%). Numbers in bold indicate the best performance and underscored ones are the second best.}
		\label{tab:sota}
		\centering
		\resizebox{0.65\linewidth}{!}{
			\begin{tabular}{clccccc}
			\toprule
			& \multicolumn{1}{c}{\multirow{2}*{Methods}} & \multicolumn{2}{c}{Market-1501} & \multicolumn{2}{c}{DukeMTMC-reID} \\ 
			\cmidrule(r){3-4} \cmidrule(r){5-6}
			& & mAP & rank-1 & mAP & rank-1  \\
			\midrule
			\parbox[t]{4mm}{\multirow{17}{*}{\rotatebox[origin=c]{90}{General reID}}}
			& PCB~\citep{sun2018beyond} 				& 77.4 & 92.3 & 66.1 & 81.7 \\
%			& FDGAN~\citep{ge2018fd} 		  		& 77.7 & 90.5 & 64.5 & 80.0 \\
			& Part-Aligned~\citep{suh2018part}   	& 79.6 & 91.7 & 69.3 & 84.4 \\
			& P$^2$-Net~\citep{guo2019beyond} 		& 85.6 & 95.2 & 73.1 & 86.5 \\
			& Top-DB-Net~\citep{quispe2021top} 		& 85.8 & 94.9 & 73.5 & 87.5 \\
			& DG-Net~\citep{zheng2019joint} 			& 86.0 & 94.8 & 74.8 & 86.6 \\
			& DRL-Net~\citep{jia2022learning} 		& 86.9 & 94.7 & 76.6 & 88.1 \\
			& MGN~\citep{wang2018learning} 			& 86.9 & 95.7 & 78.4 & 88.7 \\
			& BPBreID~\citep{somers2023body}			& 87.0 & 95.1 & 78.3 & 89.6 \\
			& ISGAN~\citep{eom2019learning} 		    & 87.1 & 95.2 & 79.5 & 90.0 \\
			& DNDM~\citep{zhao2020not} 		    		& 87.1 & 95.6 & 78.7 & 88.8 \\
			& ViT-B+DCAL~\citep{zhu2022dual}   		& 87.5 & 94.7 & 80.1 & 89.0 \\
			& DAAF~\citep{chen2022deep} 		    		& 87.9 & 95.1 & 77.9 & 87.9 \\
			& PAT~\citep{li2021diverse} 				& 88.0 & 95.4 & 78.2 & 88.8 \\
			& AdaptiveL2~\citep{ni2021adaptive} 		& 88.3 & 95.3 & 79.9 & 88.9 \\
			& RGA-SC~\citep{zhang2020relation} 		& 88.4 & \textbf{96.1}  & - & -  \\
			& SCSN~\citep{chen2020salience} 			& 88.5 & 95.7 & 79.0 & 90.1 \\
			& ISP~\citep{zhu2020identity} 			& 88.6 & 95.3 & 80.0 & 89.6 \\
			& LTReID~\citep{wang2022ltreid} 			& 89.0 & 95.9 & 80.4 & \underline{90.5} \\
			& SCAL~\citep{chen2019self} 				& 89.3 & 95.8 & 79.1 & 88.9 \\
			& CLIP-ReID~\citep{li2023clip} 				& \underline{89.6} & 95.5 & \textbf{82.5} & 90.0 \\
			\midrule
			\parbox[t]{4mm}{\multirow{10}{*}{\rotatebox[origin=c]{90}{Attribute-based reID}}}
			& UPAR~\citep{specker2023upar} 			& 40.6 & 55.4 & -    & -    \\
			& ACRN~\citep{schumann2017person} 		& 62.6 & 83.6 & 52.0 & 72.6 \\
			& APR~\citep{lin2019improving} 			& 66.9 & 87.0 & 55.6 & 73.9 \\
			& A$^3$M~\citep{han2018attribute} 		& 69.0 & 86.5 & -    & -    \\
			& UF~\citep{sun2018unified} 				& 70.1 & 87.1 & 66.7 & 80.6 \\
			& UCAD~\citep{yan2022weakening} 			& 79.5 & 92.6 & 66.7 & 80.6 \\
			& CA$^3$Net~\citep{liu2018ca3net}		& 80.0 & 93.2 & 70.2 & 84.6 \\
			& APDR~\citep{li2020attributes} 			& 80.1 & 93.1 & 69.7 & 84.3 \\
			& AANet~\citep{tay2019aanet} 			& 82.5 & 93.9 & 72.6 & 86.4 \\
			& AttKGCN~\citep{jiang2019attkgcn} 		& 85.5 & 94.4 & 77.4 & 87.8 \\
			& GPS~\citep{nguyen2021graph} 			& 87.8 & 95.2 & 78.7 & 88.2 \\
%			& AMD~\citep{chen2021explainable} 		& 88.6 & \underline{95.9} & 78.3 & 89.2 \\
			& Cerberus	           				& \textbf{89.8} & \textbf{96.1} & \underline{80.7} & \textbf{91.1} \\
			\bottomrule
			\end{tabular}
		}
	\end{table*}
	
		\subsubsection{Training} We use ResNet-50~\citep{he2016deep} trained for ImageNet classification~\citep{krizhevsky2012imagenet} as a visual encoder~(Fig.~\ref{fig:overview}). Specifically, we use the network cropped at $\texttt{conv4-1}$ as our backbone. We duplicate the remaining network, and exploit them for extracting feature maps, $\mathbf{F}^g_x$ and $\mathbf{F}^l_x$, respectively. The height, width, and channel depth of the feature maps~($H$, $W$, $D$) are set to $24$, $8$, and $2048$, respectively. The sizes of person representations~$d$ for head, upper body, lower body, identity, and carrying are $512$. The sizes of SID prototypes are also $512$, and they are initialized with the He normal initialization~\citep{he2015delving}.
				
			We train our model end-to-end for 24k iterations. We use the Adam optimizer~\citep{kingma2014adam}, where $\beta_1$ and $\beta_2$ are set to 0.9 and 0.999, respectively. Following~\citep{luo2019bag,quispe2021top,ni2021adaptive,he2020fastreid}, we adopt a warm-up and cosine annealing strategy. Specifically, the learning rate linearly increases from $3.5\times10^{-6}$ to $3.5\times10^{-4}$ for the first 2k iterations, and then decreases from the next iterations using a cosine annealing technique~\citep{loshchilov2016sgdr}. For a mini-batch, we randomly choose 16 persons, and sample 4 images for each person. We resize person images into the size of 384$\times$128, and augment them with horizontal flipping and random erasing~\citep{zhong2020random} for training. The batch-hard mining strategy~\citep{hermans2017defense} is used to set triplet pairs \{$\mathbf{f}_x^\mathrm{G}$, $\mathbf{f}_p^\mathrm{G}$, $\mathbf{f}_n^\mathrm{G}$ \} for the identification term.

	\begin{table*}[h!]
		\caption{Quantitative comparisons for PAR on Market-1501~\citep{zheng2015scalable} in terms of mA(\%). Note that methods in the first group are specially designed for PAR, while those in the second group are for attribute-based person reID. Numbers in bold indicate the best performance and underscored ones are the second best.}
		\label{tab:par_market}
		\centering
		\resizebox{\linewidth}{!}{
			\begin{tabular}{clccccccccccccc}
			\toprule
			& \multicolumn{1}{c}{\multirow{2}*{Methods}} & \multicolumn{2}{c}{Identity} & \multicolumn{3}{c}{Carrying} & \multicolumn{2}{c}{Head} & \multicolumn{2}{c}{Upper body} & \multicolumn{3}{c}{Lower body} & \multicolumn{1}{c}{\multirow{2}*{mA}} \\ 
			\cmidrule(lr){3-4} \cmidrule(lr){5-7} \cmidrule(lr){8-9} \cmidrule(lr){10-11} \cmidrule(lr){12-14} 
			& & Gender & Age & B.pack & H.bag & Bag & L.hair & Hat & \phantom{   }L.up\phantom{   } & \phantom{   }C.up\phantom{   } & L.low & C.low & S.low & \\
			\midrule
			& UPAR~\citep{specker2023upar}			& - & - & - & - & - & - & - & - & - & - & - & - & 79.5  \\
			& ARN~\citep{lin2019improving} 			& 87.5 & \underline{85.8} & 86.6 & 88.1 & 78.6 & 84.2 & \underline{97.0} & \underline{93.5} & 72.4 & \textbf{93.6} & 71.7 & \underline{93.6} & 86.0  \\
			& UF~\citep{sun2018unified} 				& 88.9 & 78.3 & \underline{93.5} & \underline{92.1} & 84.8 & \underline{97.1} & 85.5 & 67.3 & \underline{88.4} & 84.8 & 87.5 & 87.2 & 86.3  \\
			& JCM~\citep{liu2018sequence} 			& \underline{89.7} & 82.5 & \textbf{93.7} & \textbf{93.3} & \textbf{89.2} & \textbf{97.2} & 85.2 & 86.9 & 86.2 & 87.4 & \underline{92.4} & 93.1 & \underline{89.7}  \\
			& HFE~\citep{yang2020hier} 		    & \textbf{94.9} & \textbf{94.4} & 90.4 & 91.5 & \underline{85.4} & 90.5 & \textbf{97.9} & \textbf{94.0} & \textbf{94.4} & \underline{93.3} & \textbf{94.0} & \textbf{94.2} & \textbf{92.9} \\
			\midrule
			& APR~\citep{lin2019improving} 			& 88.9 & 88.6 & 84.9 & \textbf{90.4} & 76.4 & 84.4 & \underline{97.1} & \underline{93.6} & 74.0 & 93.7 & 73.8 & 92.8 & 86.6 \\
			& AANet~\citep{tay2019aanet} 		    & \underline{92.3} & 88.2 & 87.8 & \underline{89.6} & 79.7 & 86.6 & \textbf{98.0} & \textbf{94.5} & 77.1 & \underline{94.2} & 70.8 & \textbf{94.8} & 87.8 \\
			& AttKGCN~\citep{jiang2019attkgcn} 		& 89.4 & \underline{88.9} & \textbf{90.0} & 89.3 & \textbf{89.6} & \underline{90.1} & 89.5 & 89.0 & \underline{88.5} & 89.8 & \underline{90.1} & \underline{94.0} & \underline{89.8} \\
			& Cerberus			            & \textbf{94.7} & \textbf{90.8} & \underline{89.0} & 83.6 & \underline{80.4} & \textbf{91.4} & 95.2 & 90.8 & \textbf{95.9} & \textbf{94.8} & \textbf{94.1} & 92.3 & \textbf{91.1} \\
			\bottomrule
			\end{tabular}
		}
	\end{table*}
	
	\begin{table*}[h!]
		\caption{Quantitative comparisons for PAR on DukeMTMC-reID~\citep{zheng2017unlabeled} in terms of mA(\%). Note that methods in the first group are specially designed for PAR, while those in the second group are for attribute-based person reID. Numbers in bold indicate the best performance and underscored ones are the second best.}
		\label{tab:par_duke}
		\centering
		\resizebox{0.85\linewidth}{!}{
			\begin{tabular}{clcccccccccccc}
			\toprule
			& \multicolumn{1}{c}{\multirow{2}*{Methods}} & Identity & \multicolumn{3}{c}{Carrying} & \multicolumn{1}{c}{Head} & \multicolumn{2}{c}{Upper body} & \multicolumn{3}{c}{Lower body} & \multicolumn{1}{c}{\multirow{2}*{mA}} \\ 
			\cmidrule(lr){3-3} \cmidrule(lr){4-6} \cmidrule(lr){7-7} \cmidrule(lr){8-9} \cmidrule(lr){10-12} 
			& & \phantom{   } Gender \phantom{   } & B.pack & H.bag & Bag & \phantom{   } Hat \phantom{   } & L.up & C.up & C.low & C.shoes & Boots & \\
			\midrule
			& ARN~\citep{lin2019improving} 			& 82.0 & 77.5 & \underline{92.3} & 82.2 & 85.5 & 86.2 & 73.4 & 68.3 & 87.6 & 88.3 & 82.3  \\
			& UF~\citep{sun2018unified} 				& \textbf{88.9} & \textbf{93.6} & 80.1 & 83.0 & 87.0 & \textbf{91.6} & 89.6 & 83.7 & \textbf{93.9} & \underline{91.8} & 88.3  \\
			& JCM~\citep{liu2018sequence} 			& \underline{87.4} & 88.3 & 89.6 & \underline{83.3} & \textbf{89.0} & 87.9 & \underline{92.4} & \underline{87.1} & 92.9 & \textbf{92.1} & \underline{89.0}  \\
			& HFE~\citep{yang2020hier} 			& 87.0 & \underline{88.5} & \textbf{93.6} & \textbf{91.8} & \underline{88.7} & \underline{89.9} & \textbf{95.9} & \textbf{97.8} & \underline{93.8} & 90.7 & \textbf{91.8}  \\
			\midrule
			& APR~\citep{lin2019improving} 			& \underline{84.2} & \underline{75.8} & \textbf{93.4} & \underline{82.9} & \underline{87.6} & \textbf{88.4} & \underline{74.2} & \underline{69.9} & \textbf{89.7} & \underline{87.5} & \underline{83.4}  \\
			& Cerberus			            & \textbf{87.6} & \textbf{80.4} & \underline{90.1} & \textbf{88.8} & \textbf{89.8} & \underline{84.7} & \textbf{93.7} & \textbf{91.2} & \underline{88.8} & \textbf{91.1} & \textbf{88.6} \\
			\bottomrule
			\end{tabular}
		}		
	\end{table*}
					
	\begin{table}[!t]
		\caption{Quantitative comparisons for APS on Market-1501~\citep{zheng2015scalable} and DukeMTMC-reID~\citep{zheng2017unlabeled} in terms of rank-1 accuracy(\%) and mAP(\%). Note that all methods, except ours, are specialized for APS. Numbers in bold indicate the best performance and underscored ones are the second best.}
		\label{tab:sota_aps}
		\centering
		\resizebox{\linewidth}{!}{
			\begin{tabular}{lccccc}
			\toprule
			\multicolumn{1}{c}{\multirow{2}*{Methods}} & \multicolumn{2}{c}{Market-1501} & \multicolumn{2}{c}{DukeMTMC-reID} \\ 
			\cmidrule(r){2-3} \cmidrule(r){4-5}
			& mAP & rank-1 & mAP & rank-1  \\
			\midrule
			AAIPR~\citep{yin2018adversarial} 	& 20.7 & 40.3 & 15.7 & 46.6 \\
			AIHM~\citep{dong2019person} 			& 24.3 & 43.3 & \underline{17.4} & \underline{50.5} \\
			SAL~\citep{cao2020symbiotic} 		& 29.8 & 49.0 & -    & -    \\
%			SAL$\dagger$~\citep{cao2020symbiotic}& 29.4 & 44.4 & -    & -    \\
			ASMR~\citep{jeong2021asmr} 			& 31.0 & \textbf{49.6} & -    & -   \\
			UPAR~\citep{specker2023upar} 		& \textbf{32.3} & 45.0 & -    & -   \\
			\midrule
			Cerberus           			& \underline{31.7} & \underline{49.3} & \textbf{23.0} & \textbf{53.5} \\
			\bottomrule
			\end{tabular}
		}
	\end{table}

	\subsection{Results}
	\subsubsection{Quantitative results}
	
		\paragraph{\textbf{Attributed-based person reID.}} We compare in Table~\ref{tab:sota} our approach with state-of-the-art methods for attribute-based reID on Market-1501~\citep{zheng2015scalable} and DukeMTMC-reID~\citep{zheng2017unlabeled}. We also show the result of general reID methods that do not exploit person attribute labels. For fair comparison, we report the reID performance without applying any re-ranking techniques,~\emph{e.g.},~\emph{k}-reciprocal re-ranking~\citep{zhong2017re}, and exclude methods that use camera topology and timestamp information to reduce the number of possible gallery candidates~\citep{wang2019spatial,ren2021learning}. From Table~\ref{tab:sota}, we can clearly see that our model sets a new state of the art, achieving 89.8\% mAP and 96.1\% rank-1 accuracy on Market-1501 and 80.7\% mAP and 91.3\% rank-1 accuracy on DukeMTMC-reID. Note that other attribute-based reID approaches are outperformed by recent reID methods, although they use person attribute labels in addition to ID labels. For example, SCSN~\citep{chen2020salience}  performs better than GPS~\citep{nguyen2021graph} in terms of rank-1 accuracy on DukeMTMC-reID (SCSN: 90.1\% vs. GPS: 88.2\%). This might be because they overlook the conflicting goals between identifying persons and recognizing attributes, which further supports the result of our experiment in Fig.~\ref{fig:teaser}(a-b). On the contrary, we use attribute labels to guide embeddings of person representations, helping our model to learn subtle appearance variations for visually similar persons. As a result, our approach outperforms other attribute-based reID methods by a significant margin. It also performs better than general reID methods, especially on DukeMTMC-reID. Performance gains of our model compared to the second-best numbers among all reID methods are 1.2\% and 0.7\% for rank-1 accuracy and mAP, respectively. Note that we outperform Part-Aligned~\citep{suh2018part} and P$^2$-Net~\citep{guo2019beyond} that require other extra datasets for body keypoints or pixel-level semantic masks of,~\emph{e.g.},~30k images, which are very hard and expensive to obtain compared to person attribute labels.
		 
		 Last but not least, our model consists of a relatively simple network, compared to other methods that require extra networks for~\emph{e.g.},~estimating human poses ~\citep{suh2018part,guo2019beyond} or computing attention maps~\citep{li2021diverse,zhang2020relation,chen2020salience,chen2019self}. For example, our model has 19.5M fewer parameters and requires 3.94G fewer FLOPs, compared with MGN~\citep{wang2018learning}, the most widely adopted reID method~(MGN: 68.8M/14.00G vs. Ours: 49.3M/10.06G). Compared to GPS~\citep{nguyen2021graph}, the recent approach for attribute-based reID, our model uses 26.6M and 6.18G fewer parameters and FLOPs, respectively, and clearly outperforms GPS in all benchmarks.

	\begin{figure*}[t!]
	    \centering
	    \includegraphics[width=0.95\linewidth]{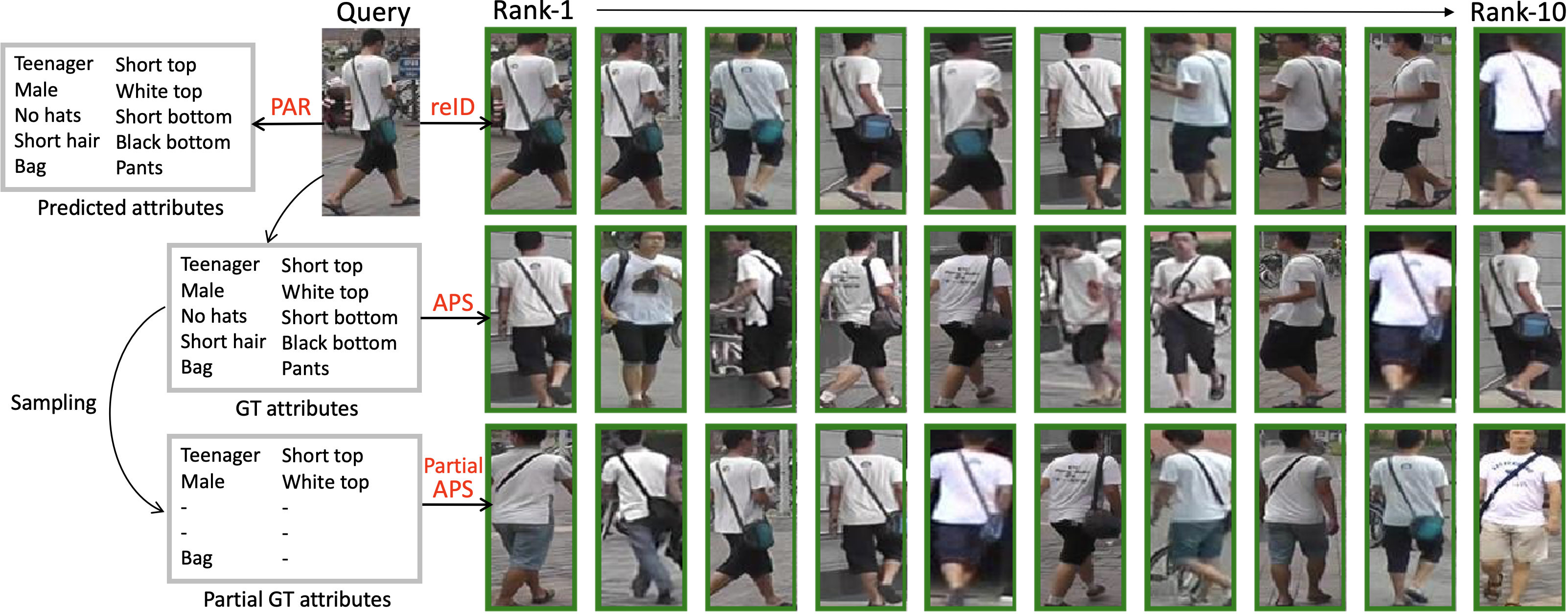}
	    \includegraphics[width=0.95\linewidth]{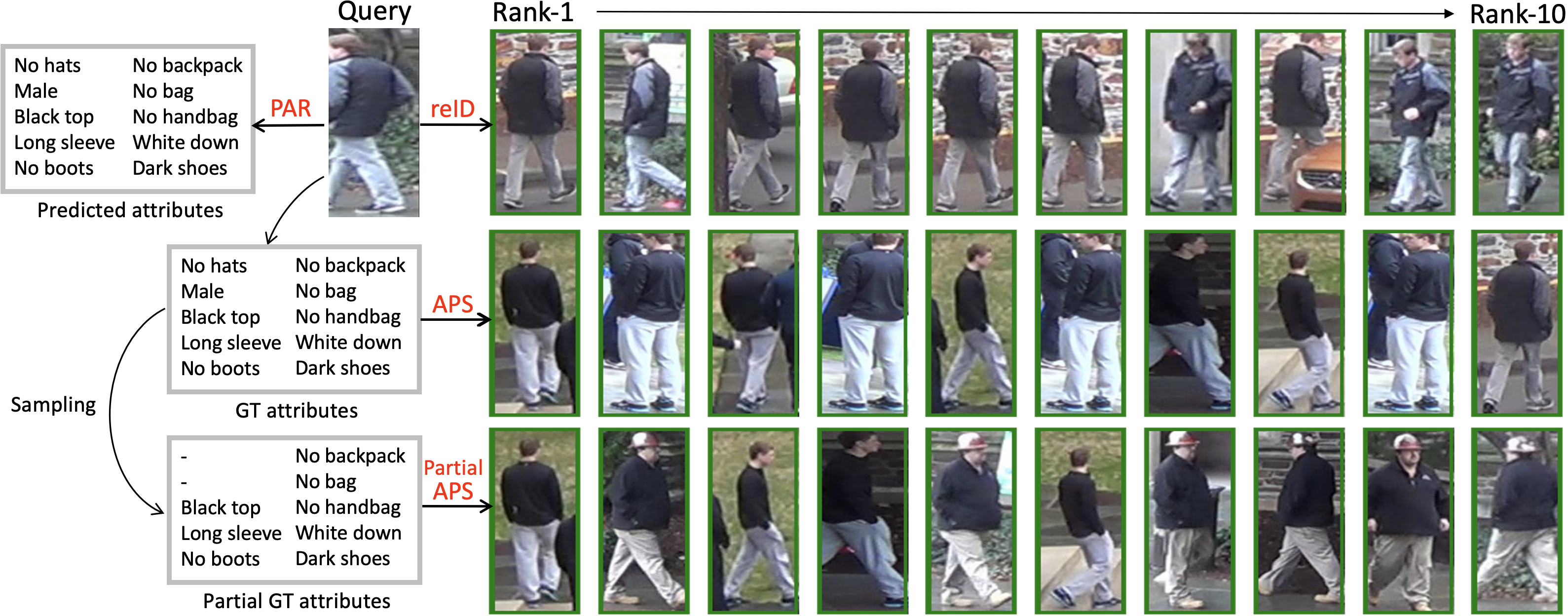}
	    \caption{Qualitative results of our model for attribute-based person reID, PAR, and APS on (top) Market-1501~\citep{zheng2015scalable} and (bottom) DukeMTMC-reID~\citep{zheng2017unlabeled}. We visualize top-10 retrieval results for reID and APS, and show predicted attribute labels for PAR. (Best viewed in color.)}
	    \label{fig:qualitative_result}
	\end{figure*}
		
		\paragraph{\textbf{PAR.}} We show in Table~\ref{tab:par_market} and Table~\ref{tab:par_duke} PAR results for each attribute on Market-1501~\citep{zheng2015scalable} and DukeMTMC-reID~\citep{zheng2017unlabeled}, respectively. The approaches in the first group~(ARN~\citep{lin2019improving}, UF~\citep{sun2018unified}, JCM~\citep{liu2018sequence}, and HFE~\citep{yang2020hier}) are specially designed for PAR, while those in the second group~(APR~\citep{lin2019improving}, AANet~\citep{tay2019aanet}, and AttKGCN~\citep{jiang2019attkgcn}) are for attribute-based reID. We can see that our model achieves the best mA among the attribute-based reID methods on the both datasets (Market-1501: 91.1\% and DukeMTMC: 88.6\%). Moreover, it even achieves comparable performance with the state of the art for PAR, HFE~\citep{yang2020hier}, without using~\emph{e.g.},~the BCE loss~\citep{lin2019improving,yang2020hier}, a localization module~\citep{tay2019aanet} or GCN~\citep{jiang2019attkgcn}, specialized for recognizing person attributes.
		
		\paragraph{\textbf{APS.}} We compare in Table~\ref{tab:sota_aps} our model with state-of-the-art APS methods on Market-1501~\citep{zheng2015scalable} and DukeMTMC-reID~\citep{zheng2017unlabeled}. Note that all methods, except ours, are specialized for APS. Instead of person images of interest, APS exploits the set of attribute labels to retrieve persons. Current APS methods~\citep{yin2018adversarial,cao2020symbiotic,jeong2021asmr,specker2023upar} consider visual-semantic alignments in a global level, and they typically leverage an adversarial learning scheme for the multi-modal alignments, which are computationally expensive in training. On the contrary, we learn multiple joint embedding spaces, where each space is specialized for learning visual-semantic alignments of particular attributes. This allows our model to match visual and semantic embeddings in a local level. We can see that our model achieves comparable or even better results than the existing APS methods. It is also worth noting that, different from AAIPR~\citep{yin2018adversarial} and ASMR~\citep{jeong2021asmr}, we do not pre-train our visual encoder~(\emph{i.e.},~feature extractor) using additional datasets for attribute classification.

		\subsubsection{Qualitative results} We show in Fig.~\ref{fig:qualitative_result} qualitative results of our model for attribute-based person reID, APS~(with selected attributes), and PAR on (top) Market-1501~\citep{zheng2015scalable} and (bottom) DukeMTMC-reID~\citep{zheng2017unlabeled}, respectively. For reID, green boxes indicate that corresponding gallery images have the same ID label as the query. We also use green boxes for APS if gallery images share the same set of attribute labels as the query. 1) \textbf{Attribute-based person reID}~(1st row): We can observe that our model retrieves images of the same person as the query, and it is robust against,~\emph{e.g.},~pose, resolution, and background variations. Also, it successfully retrieves the query person even if the backpack/bag is not clearly visible. 2) \textbf{PAR} (1st row): Our model also successfully predicts person attributes in a given query image such as gender, approximate age, or clothing/shoe color robust to~\emph{e.g.},~(top) distracting scene details and (bottom) partial occlusion. 3) \textbf{APS}~(2nd row): We assume that the image of a person of interest is not available and verbal descriptions of witnesses are the only cue for retrieving the person. Our model can find the person using the set of attribute labels as a query. We can see that it successfully finds images of the persons who have the same personal characteristics as the given attribute labels. 4) \textbf{Partial APS}~(3rd row): Our model can still find relevant candidates, even when some of the person attributes are missing, namely, information for,~\emph{e.g.},~the (top) pants or (bottom) hat is unavailable. It tries to retrieve images of the persons using available attributes only. For example, retrieved persons of APS and partial APS in Fig.~\ref{fig:qualitative_result} share the same attributes, except that,~\emph{e.g.},~(top) they wear pants of different colors/styles or (bottom) whether a hat is worn or not. To the best of our knowledge, this is the first attempt to retrieve persons of interest without access to the entire set of pre-defined attribute labels. Note also that we use a single model for three different tasks, and we do not additionally train our model for each task.
		
	\begin{table}[t]
    	\caption{Ablation studies on Market-1501~\citep{zheng2015scalable}. Numbers in bold indicate the best performance and underscored ones are the second best. AL: An alignment module.}
		\label{tab:ablation}
		\centering
		\resizebox{\linewidth}{!}{
			\begin{tabular}{p{0.2cm} p{0.35cm} p{0.35cm} p{0.3cm} c c c c c}
				\toprule
				\multicolumn{4}{c}{} & \multicolumn{2}{c}{Person reID} & \multicolumn{2}{c}{APS} & \multicolumn{1}{c}{PAR} \\
				\cmidrule(r){5-6} \cmidrule(r){7-8} \cmidrule{9-9}
				$\mathcal{L}_\mathrm{id}$ & $\mathcal{L}_\mathrm{sem}$ & $\mathcal{L}_\mathrm{reg}$ & \multicolumn{1}{c}{AL} & mAP & R-$1$ & mAP & R-$1$ & mA \\
				\midrule
				\checkmark & & & & 88.34 & 95.22 & - & - & - \\
				\checkmark & & & \checkmark & 89.47 & 95.78 & - & - & - \\
				\checkmark & \checkmark & & \checkmark & \underline{89.68} & \underline{95.90} & \underline{30.15} & \underline{48.43} & \underline{90.93} \\
				\checkmark & \checkmark & \checkmark & \checkmark & \textbf{89.83} & \textbf{96.14} & \textbf{31.66} & \textbf{49.32} & \textbf{91.13} \\
				\bottomrule
			\end{tabular}
		}
	\end{table}

	\begin{figure}
			\centering
			\renewcommand*{\thesubfigure}{}
			\subfigure[(a)]{
				\begin{minipage}[t]{0.33\linewidth}
				\raisebox{0.1\height}{
					\includegraphics[width=0.98\linewidth]{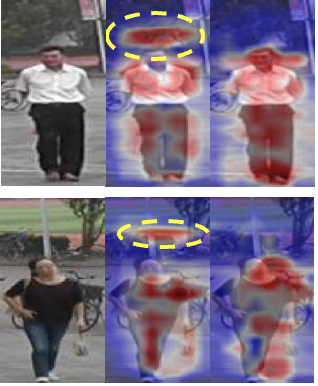}
					}
				\end{minipage}
			}
			\subfigure[(b)]{
				\begin{minipage}[t]{0.57\linewidth}
					\begin{overpic}[width=\linewidth]{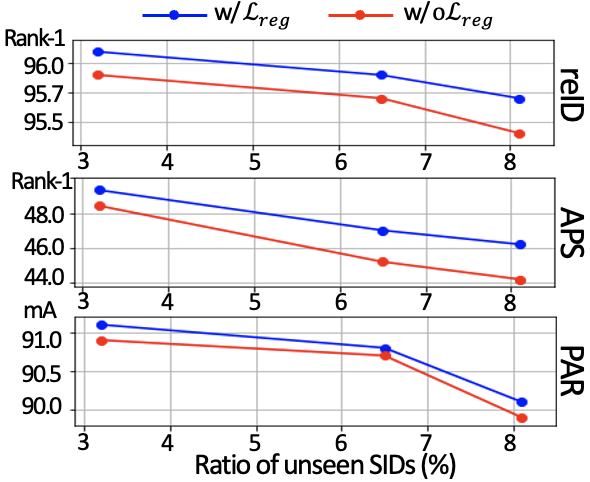}
					\end{overpic}
				\end{minipage}
			}
		\caption{(a) An input image (left), and a heat map, obtained by our model trained without (middle) and with (right) the alignment module. (b) Quantitative comparisons for the regularization term w.r.t the number of unseen SIDs. We obtain both results on Market-1501~\citep{zheng2015scalable}.}
	    \label{fig:disscussion1}
	\end{figure}

		\subsection{Discussion} \label{subsec:discussion}

		\subsubsection{Ablation study} We show an ablation study of our model on Market-1501~\citep{zheng2015scalable} in Table~\ref{tab:ablation}. From the first and second rows, we can see that our alignment module boosts the reID performance. We visualize in Fig.~\ref{fig:disscussion1}(a) input person images (left), and heat maps $\mathbf{H}_x$, obtained from our model, without (middle) and with (right) the alignment module. Without the alignment module, our model heavily focuses on distracting scene details,~\emph{e.g.},~trees or bushes in background, which causes person representations to encode such distracting details. The alignment module reduces this problem, and allows our model to extract person representations focusing on human body parts. The second and third rows show the effectiveness of the semantic guidance term. Note that we use SID prototypes for guiding embeddings of person representations only at training time, and do not exploit them for the reID task during evaluation. Even though we do not use any additional parameters, the semantic guidance term can clearly improve the reID performance. Furthermore, when we exploit learned SID prototypes with negligible memory and computational costs~(\emph{i.e.},~0.05M parameters and 1.03G FLOPs), our framework can perform APS and PAR without additional training. Lastly, the third and last rows show the effect of the regulation term. For the Market-1501 dataset, about 3\% of SIDs of test samples are unseen at training time. This suggests that, without our regularization term, prototypes of unseen SIDs could not be learned, degrading the performance of our model. Using the regularization term, we can train the prototypes of unseen SIDs based on the relationship between other prototypes, improving the reID performance. To further demonstrate the effectiveness of our regularization term, we randomly sample SIDs from the test samples, and exclude the images of persons belonging to the sampled SIDs from training. That is, the number of unseen SIDs is manually adjusted during evaluation. We then compare the performance of our model trained with and without the regularization term. We report rank-1 for reID and APS, and report mA for PAR task in Fig.~\ref{fig:disscussion1}(b). We can clearly see that the model trained with the proposed regularization term consistently outperforms the other one, demonstrating the effectiveness on enhancing the generalization ability of our model.	
		
	\begin{table}[t]
    	\caption{Analysis of initializing SID Prototypes on Market-1501~\citep{zheng2015scalable}. Numbers in bold indicate the best performance and underscored ones are the second best.}
		\label{tab:sid_init}
		\centering
		\resizebox{\linewidth}{!}{
			\begin{tabular}{p{3.2cm} c c c c c}
				\toprule
				\multicolumn{1}{c}{} & \multicolumn{2}{c}{Person reID} & \multicolumn{2}{c}{APS} & \multicolumn{1}{c}{PAR} \\
				\cmidrule(r){2-3} \cmidrule(r){4-5} \cmidrule{6-6}
				Initialization & mAP & R-$1$ & mAP & R-$1$ & mA \\
				\midrule
				He normal (ours) & \textbf{89.83} & \textbf{96.14} & \textbf{31.66} & \textbf{49.32} & \textbf{91.13} \\
				\midrule
				He uniform & 89.72 & 95.93 & 30.12 & 47.74 & 90.03 \\
				Xavier normal & 89.39 & 95.81 & \underline{30.98} & 47.51 & 90.41 \\
				Xavier uniform  & 89.69 & 95.62 & 30.36 & 47.57 & 90.15 \\
				Rand normal(std=0.01) & 89.48 & \underline{96.01} & 29.24 & 47.05 & 90.24 \\
				Rand normal(std=0.1) & 89.68 & 95.23 & 30.67 & 47.54 & 90.85 \\
				Rand normal(std=1) & 89.72 & 95.83 & \underline{30.98} & \underline{48.96} & \underline{91.09} \\
				Rand normal(std=10) & \underline{89.74} & 95.87 & 30.97 & 47.42 & 90.86 \\
				\bottomrule
			\end{tabular}
		}
	\end{table}

	\begin{figure}[t]
	    \centering
	    \includegraphics[width=0.9\linewidth]{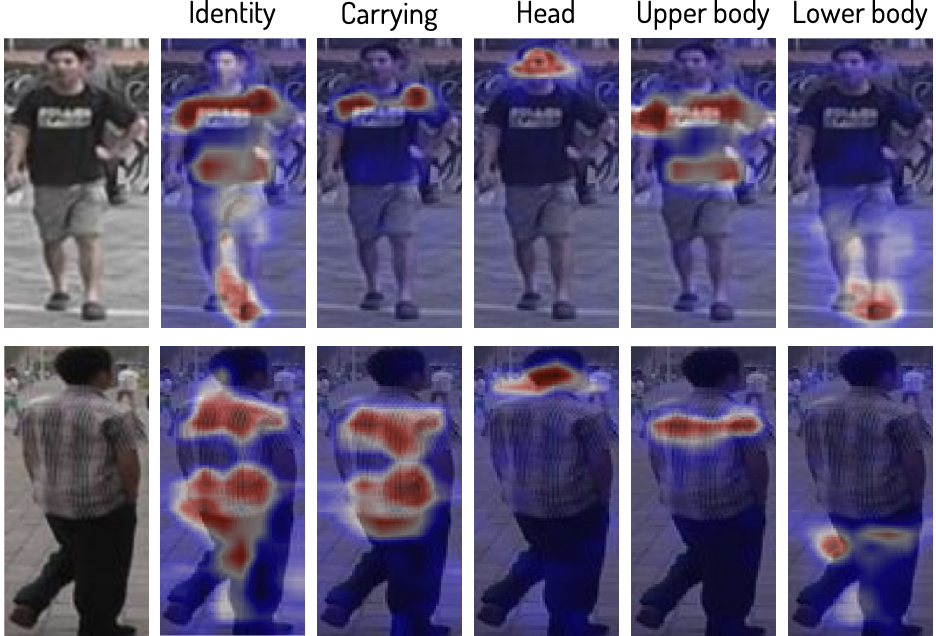}
	    \caption{Visualization of attention maps for partial person representations  on Market-$1501$~\citep{zheng2015scalable}.~(Best viewed in color.)}
	    \label{fig:part_attention}
	\end{figure}

			SID prototypes are learnable parameters trained with the visual encoder end-to-end. To demonstrate the consistent performance of our model regardless of initialization methods for SID prototypes, we compare models trained with different initialization methods for SID prototypes in Table~\ref{tab:sid_init}. We can see that neither the He uniform nor the Xavier initialization~\citep{glorot2010understanding} significantly impact on the performance of our model. Similarly, initializing SID prototypes with random normalization using varying standard deviations also has a marginal effect on the final performance. This shows the robustness of our method to initialization methods for SID prototypes.
			
	\renewcommand{\arraystretch}{1.3}
	\begin{table}[t]
	\caption{Analysis on the effect of boundary margins on Market-$1501$~\citep{zheng2015scalable} in terms of rank-1 accuracy(\%) and mAP(\%). Numbers in bold indicate the best performance and underscored ones are the second best.}
	\label{tab:boundary_margin}
	\centering
	\resizebox{0.5\linewidth}{!}{
	\begin{tabular}{l c c}
	\toprule
	\multicolumn{1}{c}{} & mAP & R-$1$ \\
	\midrule
	$m^\mathrm{G}_\mathrm{g}=0$ 	& 88.92 & 95.64 \\
	$m^\mathrm{G}_\mathrm{g}=0.6$ 	& \underline{89.60} & \underline{95.81} \\
	Ours 	& \textbf{89.83} & \textbf{96.14} \\
	\bottomrule
	\end{tabular}
	}
	\end{table}
	
	\begin{figure}[t]
	    \centering
	    \includegraphics[width=\linewidth]{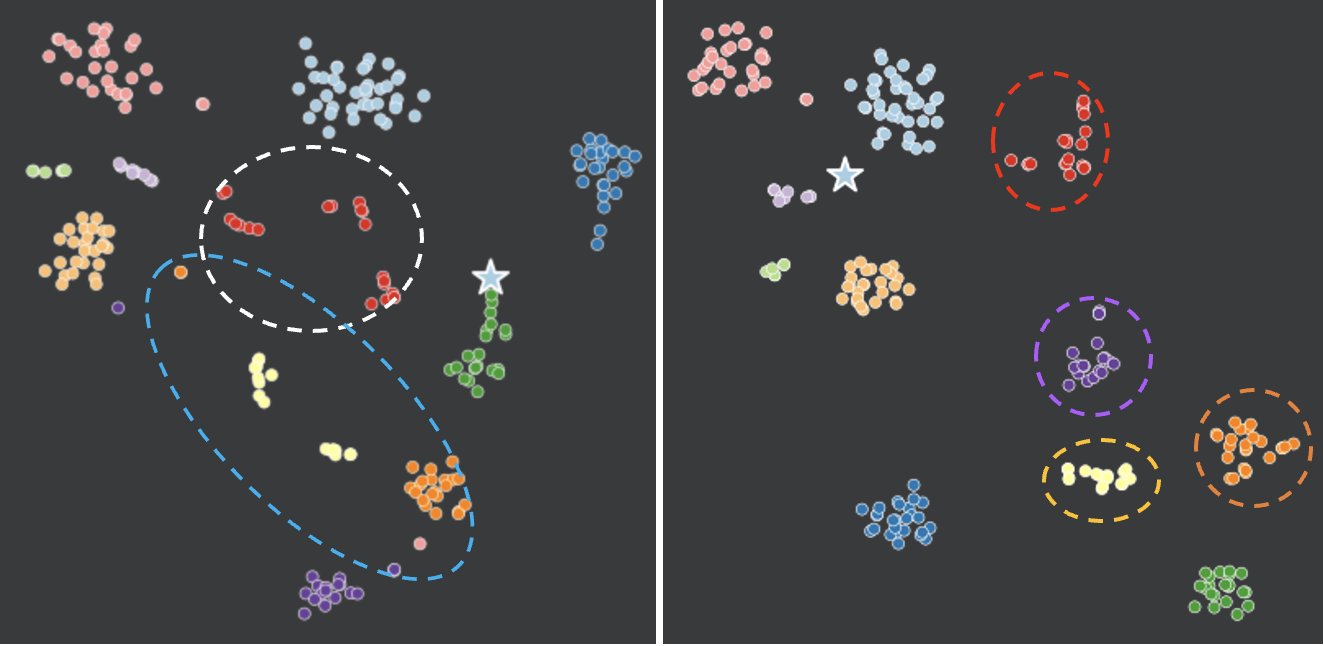}
	    \caption{t-SNE visualization of person representations: (left) a zero margin~($m^\mathrm{G}_\mathrm{g}=0$) and (right) an adaptive margin. We randomly sample 11 identities from the test split of Market-$1501$~\citep{zheng2015scalable}., and assign the same color for the representations of persons with the same identity.  
	    }
	    \label{fig:disscussion2}
	\end{figure}

		\subsubsection{Partial representation} We show visual attention maps in Fig.~\ref{fig:part_attention} to illustrate which parts of the image each partial representation encodes. The leftmost image is the original image, followed by visual attention maps for partial representations of identity, carrying, head, upper body, and lower body. For the identity representation, background regions remain inactive while the entire body, including the face and hair, is strongly activated. In the case of carrying, our model attends on the shoulder strongly, when the backpack is clear visible (Fig.~\ref{fig:part_attention}(top)). Otherwise, for the absence of carrying (Fig.~\ref{fig:part_attention}(bottom)), it is highly activated around the regions carrying objects are likely to be, such as shoulders, hands, and back. We can also see that head, upper body, and lower body representations are activated on personal traits of the respective parts of the image. For example, the upper body representation shows strong activations on logos (Fig.~\ref{fig:part_attention}(top)) or patterns (Fig.~\ref{fig:part_attention}(bottom)) on the T-shirt, which can help identify the person.

	\begin{figure*}
	\centering
	\renewcommand*{\thesubfigure}{}
	\subfigure[(a) $\sigma$]{
	\begin{minipage}[t]{0.184\linewidth}
	\raisebox{0.\height}{
	\includegraphics[width=\linewidth]{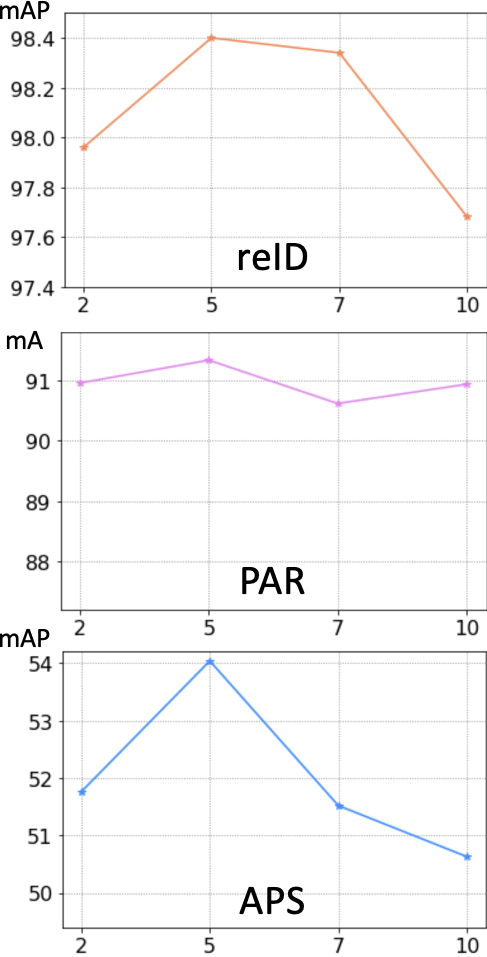}
	}
	\end{minipage}
	}
	\subfigure[(b) $\lambda_{sem}$]{
	\begin{minipage}[t]{0.185\linewidth}
	\raisebox{0.\height}{
	\includegraphics[width=\linewidth]{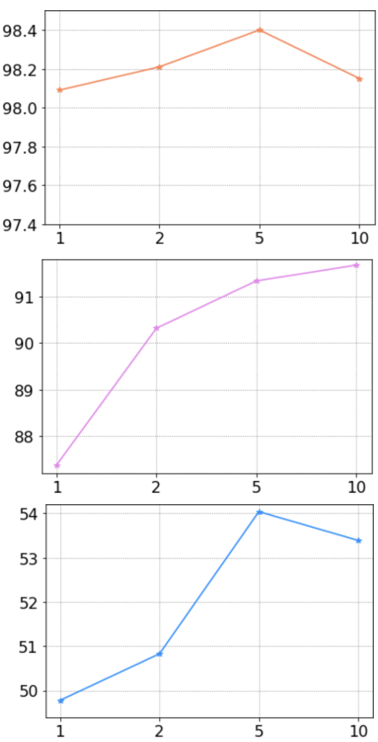}
	}
	\end{minipage}
	}
	\subfigure[(c) $\lambda_{reg}$]{
	\begin{minipage}[t]{0.187\linewidth}
	\raisebox{0.\height}{
	\includegraphics[width=\linewidth]{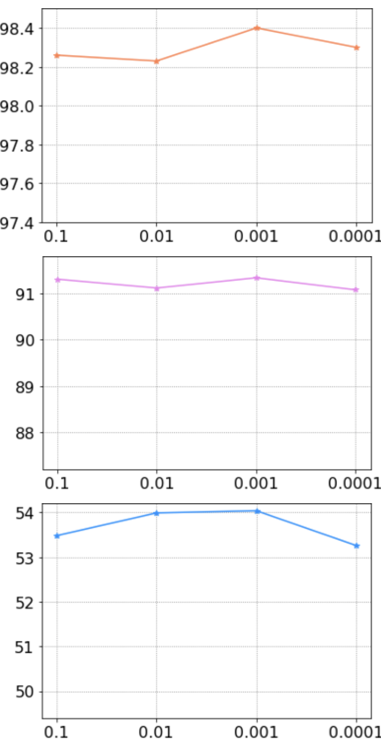}
	}
	\end{minipage}
	}
	\subfigure[(d) $\alpha$]{
	\begin{minipage}[t]{0.185\linewidth}
	\raisebox{0.\height}{
	\includegraphics[width=\linewidth]{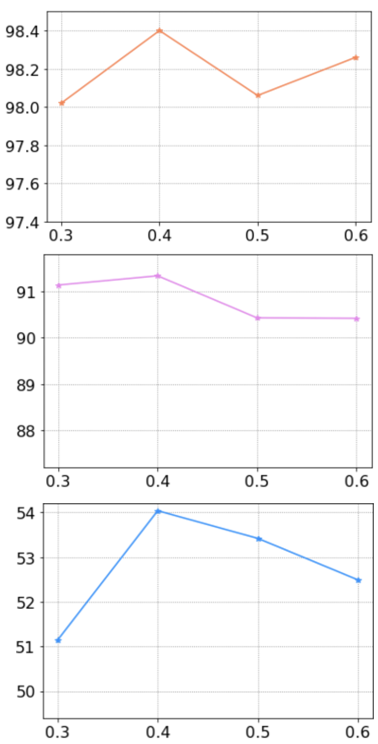}
	}
	\end{minipage}
	}
	\subfigure[(e) $\beta$]{
	\begin{minipage}[t]{0.184\linewidth}
	\raisebox{0.\height}{
	\includegraphics[width=\linewidth]{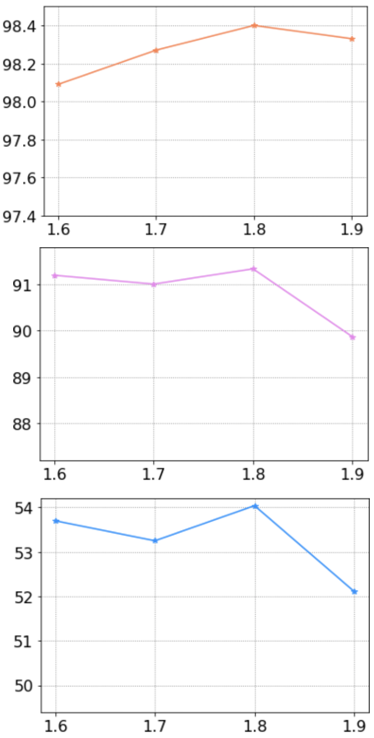}
	}
	\end{minipage}
	}
	\caption{Sensitivity analysis of hyperparameters on Market-$1501$~\citep{zheng2015scalable}. The top line displays the reID performance, while the middle and bottom lines show the PAR and APS results, respectively. The mAP is used for person reID and APS, and the mA is reported for PAR. We perform a grid search for (a) the threshold value $\sigma$ in the alignment module, (b-c) loss balance parameters, $\lambda_{sem}$ and $\lambda_{reg}$, and (d-e) the parameters, $\alpha$ and $\beta$, that control the boundary margins.}
	\label{fig:hyperparam}
	\end{figure*}

		\subsubsection{Boundary margin} In Table~\ref{tab:boundary_margin}, we evaluate the effect of the boundary margin, $m^\mathrm{G}_\mathrm{g}$ in Eq.~\eqref{eq:l_sem}, on person reID. The semantic guidance term encourages our person representations to encode specific personal traits by pulling the representations closer to corresponding SID prototypes. When the distance between them is smaller than the boundary margin, the representations are mainly guided by the identification term, which encourages the model to discriminate visually similar persons. However, when the boundary margin is set to zero, the semantic guidance term continually forces the person representations to be similar to the prototypes. This forces the person representations to encode visual commonness between persons with the same attributes, interfering with learning the visual differences between them, and consequently may cause the conflicting goal problem shown in Fig.~\ref{fig:teaser}. Thereby, the reID performance is significantly reduced as in the first row of Table~\ref{tab:boundary_margin}.

			To further support this observation, we visualize the t-SNE~\citep{maaten2008visualizing} embeddings of person representations of ten different persons who belong to the same SID (\emph{e.g.},~they are wearing the same color of upper clothing with the same sleeve length) in Fig.~\ref{fig:disscussion2}. The representations of persons with the same identity are assigned the same color. With a boundary margin of zero, the representations of the same person are dispersed in the learned embedding space, and they may even be mapped close to representations of different IDs (Fig.~\ref{fig:disscussion2}(left)). On the other hand, our model forms compact clusters that match ID labels when the boundary margin is used (Fig.~\ref{fig:disscussion2}(right)), suggesting that the margin helps to better differentiate the subtle appearance differences between the persons sharing the same attribute labels.
			
			We adaptively adjust the boundary margin for a particular SID based on the number of persons in the SID, as in Eq.~\eqref{eq:boundary_margin}. To demonstrate the effectiveness of the adaptive margin, we compare our model trained with a fixed margin, set to the average value of the adaptive ones~(\emph{i.e.},~$m^\mathrm{G}_\mathrm{g}=0.6$). The results in the second and last rows of Table~\ref{tab:boundary_margin} clearly show that using the adaptive margin performs better. This indicates that when more persons belong to the same SID, it is important to focus on distinguishing the subtle differences.	
	
		\subsubsection{Hyperparameters} 
		
		To determine hyperparameters, we divide the training split of Market-1501~\citep{zheng2015scalable} into two subsets: a training subset with 651 IDs and a validation subset with 100 IDs. We randomly sample 160 images from the validation subset to serve as queries, and use the rest as the gallery set. We show in Fig.~\ref{fig:hyperparam} mAP(\%) for person reID and APS, and mA(\%) for PAR, according to the hyperparameters. We perform a grid search over $\{2, 5, 7, 10 \}$ to set $\sigma$~(Fig.~\ref{fig:hyperparam}(a)). For the balance parameters, we set $\lambda_{id}$ to 1 for a reference point, and use a grid search to set others, $\lambda_{sem} \in \{1, 2, 5, 10 \}$~(Fig.~\ref{fig:hyperparam}(b)) and $\lambda_{reg} \in \{0.1, 0.01, 0.001, 0.0001 \}$~(Fig.~\ref{fig:hyperparam}(c)). For $\alpha$ and $\beta$, we search over $\{0.3, 0.4, 0.5, 0.6 \}$ and $\{1.6, 1.7, 1.8, 1.9 \}$, respectively~(Fig.~\ref{fig:hyperparam}(d) and Fig.~\ref{fig:hyperparam}(e)). 		The best hyperparameter values, $\sigma$ = 5, $\lambda_{id}$ = 1, $\lambda_{sem}$ = 5, $\lambda_{reg}$ = 0.001, $\alpha$ = 0.4, and $\beta$ = 1.8, are used to train our models on both Market-1501~\citep{zheng2015scalable} and DukeMTMC-reID~\citep{zheng2017unlabeled} with the same parameters.

\section{Conclusion}
We have presented a novel framework for attribute-based person reID that leverages person attribute labels to guide the embedding of person representations. To achieve this, we have defined SIDs by combining attribute labels, and learned corresponding prototypical features. We have also introduced a semantic guidance loss to align person representations with the corresponding prototypes, thereby promoting the encoding of specific personal traits in the representations. Additionally, we have proposed a regularization method that enables estimating prototypes for unseen SIDs. We have demonstrated that our framework outperforms existing attribute-based re-identification methods on standard benchmarks, and it can handle both PAR and APS effectively without bells and whistles.

\section*{Acknowledgments}
This work was partly supported by IITP grants funded by the Korea government (MSIT) (No.RS-2022-00143524, Development of Fundamental Technology and Integrated Solution for Next-Generation Automatic Artificial Intelligence System, No.2022-0-00124, Development of Artificial Intelligence Technology for Self-Improving Competency-Aware Learning Capabilities).


%\bibliographystyle{cas-model2-names}
\bibliographystyle{apalike2}
\bibliography{egbib} % name your BibTeX data base

% % Biography
% % Here goes the biography details.

% % Here goes the biography details.

\end{document}